\pdfoutput=1

\documentclass[11pt]{article}

\usepackage[final]{acl}

\usepackage{times}
\usepackage{latexsym}

\usepackage[T1]{fontenc}

\usepackage[utf8]{inputenc}

\usepackage{microtype}

\usepackage{inconsolata}

\usepackage{microtype}
\usepackage{inconsolata}
\usepackage{array}
\usepackage{tabularx}
\usepackage{hyperref}
\usepackage{mathtools}
\usepackage{todonotes}
\usepackage{booktabs}
\usepackage{soul} 
\usepackage{tabu}
\usepackage{nicematrix}
\usepackage{tikz}
\usepackage{ulem}
\usepackage{cancel}

\usepackage{times}
\usepackage{latexsym}
\usepackage{multirow}
\usepackage{sidecap}
\usepackage{amsmath}
\usepackage{amssymb}
\usepackage{amsfonts}
\usepackage{verbatim} 
\usepackage{textcomp}
\usepackage{url}
\usepackage{pifont}
\usepackage{eqparbox}
\usepackage{comment}
\usepackage{comment}
\usepackage{booktabs}
\usepackage{enumitem}
\usepackage{caption}
\usepackage{subcaption}
\usepackage{bm}
\usepackage{MnSymbol,bbding,pifont}
\usepackage[most]{tcolorbox} 
\usepackage{cleveref}
\usepackage{makecell}
\usepackage{xcolor}  
\usepackage{graphicx}  

\newtcbtheorem[auto counter, number within=section, 
crefname={example}{Example},
Crefname={Example}{Example}]
{exmp}{Exam\smash{p}le} 
{colback=pale_red!5, colframe=DarkGreen, left=.02in, right=.02in,bottom=.02in, top=.02in}{exmp}

\definecolor{lightergray}{RGB}{230,230,230}
\definecolor{DarkRed}{RGB}{130,25,0}
\definecolor{PurpleRed}{RGB}{204,0,102}
\definecolor{DarkGreen}{RGB}{30,130,30}
\definecolor{DarkBlue}{RGB}{0,0,250}
\definecolor{DarkYellow}{RGB}{255,128,0}
\definecolor{light-gray}{gray}{0.95}
\definecolor{lightgreen}{RGB}{231,255,219}
\definecolor{lightred}{RGB}{252,231,234}
\definecolor{lightyellow}{RGB}{250,253,191}
\definecolor{lightpurple}{RGB}{229,204,255}
\definecolor{lightblue}{RGB}{229,246,254}
\definecolor{value-modification}{RGB}{250, 217, 86}
\definecolor{digit-expansion}{RGB}{216, 194, 104}
\definecolor{integer-decimal-fraction}{RGB}{240, 133, 51}
\definecolor{semantic-paraphrasing}{RGB}{85, 157, 63}
\definecolor{complexity-increasing}{RGB}{58, 120, 175}
\definecolor{question-transformation}{RGB}{174, 205, 225}
\definecolor{interference-injection}{RGB}{255,204,229}
\definecolor{remove-constrain}{RGB}{204,204,255}
\definecolor{myGreen}{RGB}{127,210,85}
\definecolor{myOrange}{RGB}{242,154,66}
\definecolor{myYellow}{RGB}{247,223,65}
\definecolor{myRed}{RGB}{232,80,43}
\definecolor{myViolet}{RGB}{162,57,102}
\definecolor{myBlue}{HTML}{4686f3}
\definecolor{myYellowv2}{HTML}{E6C802}
\definecolor{myOrangev2}{HTML}{ED8E55}
\definecolor{MyGreenv2}{HTML}{009B55}
\definecolor{MyRedv2}{HTML}{c22f2f}

\newcommand\encircle[2][]{\tikz[overlay]\node[fill=blue!20,inner sep=2pt, anchor=text, rectangle, rounded corners=1.5mm,#1] {#2};\phantom{#2}}

\definecolor{pale_green}{rgb}{0.55,0.75,0.60}
\definecolor{pale_red}{rgb}{0.90,0.61,0.58}
\definecolor{pale_yellow}{rgb}{0.95,0.92,0.72}

\usepackage{xspace}

\definecolor{Gray}{gray}{0.94}

\title{A Survey of Mathematical Reasoning in the Era of \\Multimodal Large Language Model: Benchmark, Method \& Challenges}

\author{%
  Yibo Yan$^{1,2}$,
  Jiamin Su$^{1}$,
  Jianxiang He$^{1}$,
  Fangteng Fu$^{1}$,
  Xu Zheng$^{1,2}$,\\
  \textbf{Yuanhuiyi Lyu}$^{1,2}$, 
  \textbf{Kun Wang}$^{3}$,
  \textbf{Shen Wang}$^{4}$,
  \textbf{Qingsong Wen}$^{4}$,
  \textbf{Xuming~Hu}$^{1,2,}$\thanks{~Corresponding Author} \\
  \fontsize{9.0pt}{\baselineskip}\selectfont $^{1}$ The Hong Kong University of Science and Technology (Guangzhou),  
  \fontsize{9.0pt}{\baselineskip}\selectfont $^{2}$ The Hong Kong University of Science and Technology,\\
  \fontsize{9.0pt}{\baselineskip}\selectfont $^{3}$ Nanyang Technological University,
  \fontsize{9.0pt}{\baselineskip}\selectfont $^{4}$ Squirrel Ai Learning \\
   \fontsize{9.0pt}{\baselineskip}\selectfont\texttt{yanyibo70@gmail.com}, \texttt{xuminghu@hkust-gz.edu.cn}
}

\begin{document}
\maketitle
\begin{abstract}
Mathematical reasoning, a core aspect of human cognition, is vital across many domains, from educational problem-solving to scientific advancements. As artificial general intelligence (AGI) progresses, integrating large language models (LLMs) with mathematical reasoning tasks is becoming increasingly significant. This survey provides the \textbf{first comprehensive analysis of mathematical reasoning in the era of multimodal large language models (MLLMs)}. We review over 200 studies published since 2021, and examine the state-of-the-art developments in Math-LLMs, with a focus on multimodal settings. We categorize the field into three dimensions: \textit{benchmarks}, \textit{methodologies}, and \textit{challenges}. In particular, we explore multimodal mathematical reasoning pipeline, as well as the role of (M)LLMs and the associated methodologies. Finally, we identify seven major challenges hindering the realization of AGI in this domain, offering insights into the future direction for enhancing multimodal reasoning capabilities. This survey serves as a critical resource for the research community in advancing the capabilities of LLMs to tackle complex multimodal reasoning tasks.
\end{abstract}

\section{Introduction}

Mathematical reasoning is a critical aspect of human cognitive ability, involving the process of deriving conclusions from a set of premises through logical and systematic thinking \cite{jonsson2022creative,yu2024natural}. It plays an essential role in a wide range of applications, from problem-solving in education to advanced scientific discoveries. As artificial general intelligence (AGI) continues to advance \cite{zhong2024evaluation}, the integration of large language models (LLMs) with mathematical reasoning tasks becomes increasingly significant. These models, with their impressive capabilities in language understanding, have the potential to simulate complex reasoning processes that were once thought to be inherently human. In recent years, both academia and industry have placed increasing emphasis on this direction \cite{wang2024large,xu2024large,lu2022survey,yan2025position}. 

\begin{figure}[!t]
    \centering
    \includegraphics[width=\linewidth,scale=1.00]{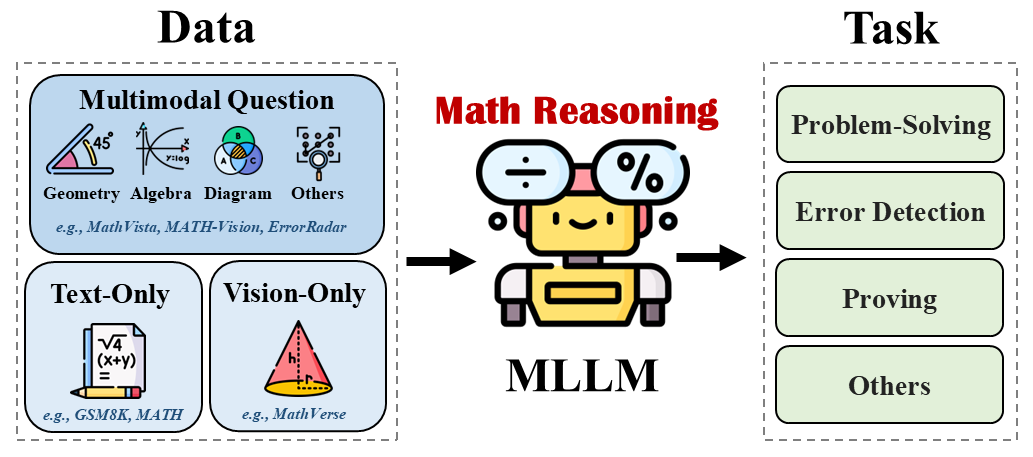}
    \caption{The illustration of our research scope (\textit{i.e.,}  investigating the MLLM's math reasoning capability).}
    \label{fig:big_picture}
\end{figure}

\begin{table}[!t] 
    \centering\small
    \resizebox{\columnwidth}{!}{
    \begin{tabular}{lccccc}  
    \toprule
    \textbf{Survey}  & \textbf{Venue \& Year} &  \textbf{Scope} & \textbf{Multimodal} & \textbf{LLM} \\
    \midrule  
    \cite{o2015language} & JMB'15 & MM4Math & \ding{52} &  \\
    \cite{hegedus2015foundations} & IRME'15 & MM4Math & \ding{52} &  \\
    \cite{lu2022survey} & ACL'22 & DL4Math & &  \\
    \cite{li2023adapting} & arXiv'23 & LLM4Edu &  & \ding{52} \\
    \cite{liu2023mathematical} & arXiv'23 & LLM4Edu &  & \ding{52} \\
    \cite{li2024survey} & COLM'24 & DL4TP &  &  \\
    \cite{ahn2024large} & EACL'24 & LLM4Math &  & \ding{52} \\
    \cite{xu2024large} & IJMLC'24 & LLM4Edu &  & \ding{52} \\
    \cite{wang2024large} & arXiv'24 & LLM4Edu &  & \ding{52}  \\
    \midrule
    \textbf{Ours} & \textbf{ACL'25} & \textbf{MLLM4Math} & \ding{52} & \ding{52}  \\
    \bottomrule    
    \end{tabular}}
    \caption{Comparisons between relevant surveys \& ours.} 
    \label{tab:survey_comparison}
\end{table}

\begin{figure*}[t!]
    \centering
    \includegraphics[width=\linewidth]{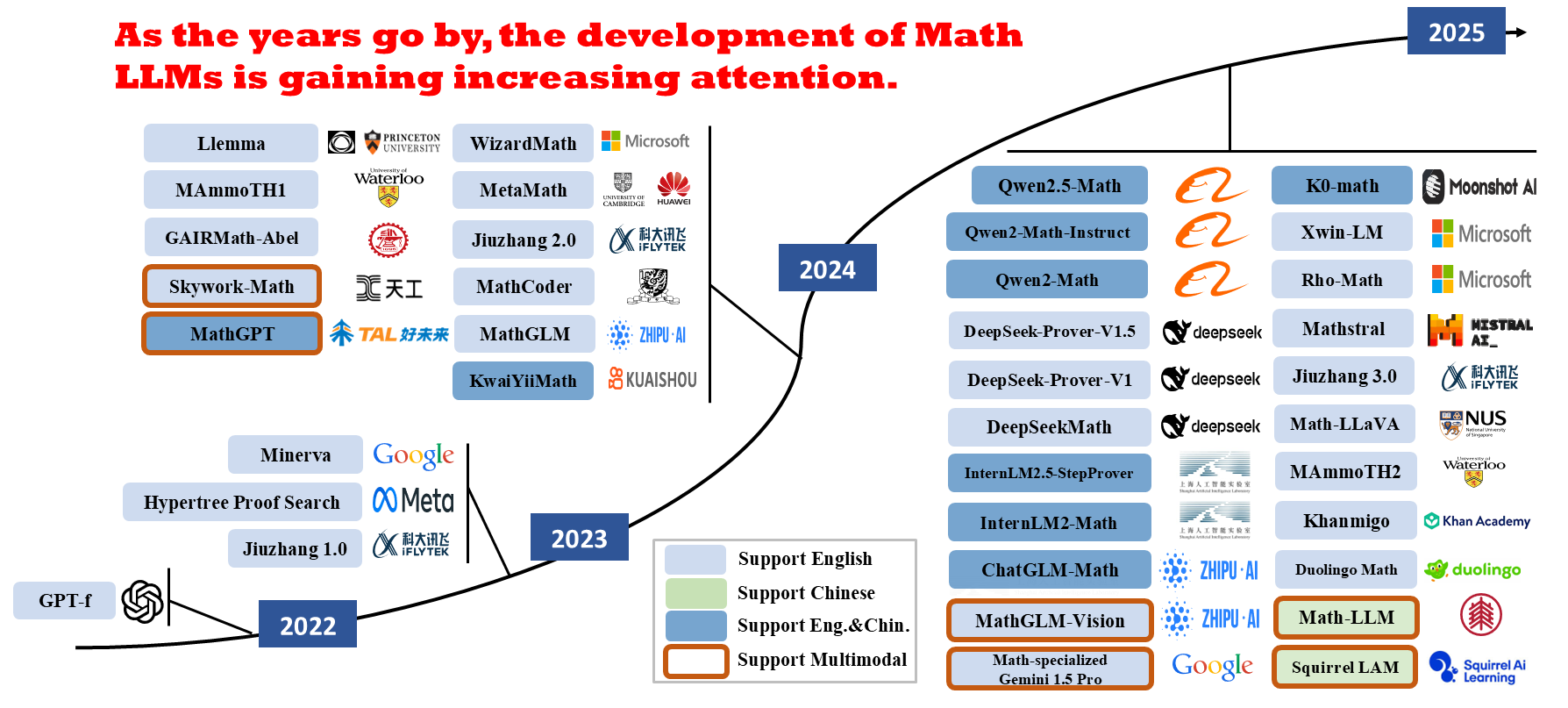}
    \caption{The release timeline of Math-LLMs in recent years.} 
    \label{fig:math_llm_trend}
\end{figure*}

The inputs for mathematical reasoning tasks are diverse, extending beyond traditional text-only to multimodal settings, as illustrated in Figure \ref{fig:big_picture}. Mathematical problems often involve not only textual information but also visual elements, such as diagrams, graphs, or equations, which provide essential context for solving the problem \cite{wang2024exploring,yin2024survey}. In the past year, multimodal mathematical reasoning has emerged as a key focus for multimodal large language models (MLLMs) \cite{zhang2024mm,bai2024survey,wu2023multimodal}. This shift is driven by the recognition that reasoning tasks in fields like mathematics require models capable of integrating and processing multiple modalities simultaneously to achieve human-like performance. However, multimodal mathematical reasoning poses significant challenges due to the complex interaction between different modalities, the need for deep semantic understanding, and the importance of context preservation across modalities \cite{liang2024quantifying,song2023bridge,fu2025blink}. These challenges are central to the realization of AGI, where models must integrate diverse forms of knowledge seamlessly to perform sophisticated reasoning tasks.

\textbf{Math-LLM Progress}. Figure \ref{fig:math_llm_trend} illustrates that, driven by the rapid development of LLMs since 2021, the number of math-specific LLMs (Math-LLMs) has grown steadily, alongside enhanced support for multilingual and multimodal capabilities (More details in Appendix \ref{app:math_llm}). The landscape was marked by the introduction of models like GPT-f \cite{polu2020generative} and Minerva \cite{lewkowycz2022solving}, with Hypertree Proof Search \cite{lample2022hypertree} and Jiuzhang 1.0 \cite{zhao2022jiuzhang} highlighting advancements in theorem proving and mathematical question understanding capabilities, respectively. Year 2023 saw a surge in diversity and specialization, alongside multimodal support from models like Skywork-Math \cite{zeng2024skywork}. In year 2024, there was a clear focus on enhancing mathematical instruction (\textit{e.g.}, Qwen2.5-Math \cite{yang2024qwen25}) and proof (\textit{e.g.}, DeepSeek-Proof \cite{xin2024deepseekp}) capabilities. The year also witnessed the emergence of Math-LLMs with a vision component, such as MathGLM-Vision \cite{yang2024mathglm}.

\textbf{Scope}. Previous surveys have not fully captured the progress and challenges of mathematical reasoning in the age of MLLMs. As indicated in Table \ref{tab:survey_comparison}, some works have concentrated on the application of deep learning techniques to mathematical reasoning \cite{lu2022survey} or specific domains such as theorem proving \cite{li2024survey}, but they have overlooked the rapid advancements brought about by the rise of LLMs. Others have broadened the scope to include the role of LLMs in education \cite{wang2024large,xu2024large,li2023adapting} or mathematical fields \cite{ahn2024large,liu2023mathematical}, but have failed to explore the development and challenges of mathematical reasoning in multimodal settings in depth. Therefore, this survey aims to fill this gap by providing the \textbf{first-ever comprehensive analysis of the current state of mathematical reasoning in the era of MLLMs}, focusing on three key dimensions: \textit{benchmark}, \textit{methodology}, and \textit{challenges}. 

\textbf{Structure}. In this paper, we survey over 200 publications from the AI community since 2021 related to (M)LLM-based mathematical reasoning, and summarize the progress of Math-LLMs. We first approach the field from the benchmark perspective, analyzing the LLM-based mathematical reasoning task through four key aspects: basic focus, task, evaluation, and training data (Section \ref{sec:bench}). Subsequently, we explore the roles that (M)LLMs play in mathematical reasoning, categorizing them as reasoners, enhancers, and planners (Section \ref{sec:method}). Finally, we identify seven core challenges that the mathematical reasoning faces in the era of MLLMs (Section \ref{sec:challenge}). This survey aims to provide the community with comprehensive insights for advancing multimodal reasoning capabilities of LLMs.

\section{Benchmark Perspective}
\label{sec:bench}
\subsection{Overview}
Benchmarking for mathematical reasoning plays a crucial role in advancing LLM research, as it provides standardized, reproducible pipeline for assessing the performance on reasoning tasks. While previous benchmarks such as GSM8K \cite{cobbe2021training} and MathQA \cite{amini2019mathqa} were instrumental in the pre-LLM era, our scope is centered on those relevant to (M)LLMs. In this section, we present a comprehensive analysis of recent benchmarks for mathematical reasoning in the context of (M)LLMs (Shown in Table \ref{tab:eval_datasets} from Appendix \ref{app:benchmarks}). The section is organized into four subsections: Basic Focus (Sec.\ref{sec:dataset}), Tasks (Sec.\ref{sec:task}), Evaluation (Sec.\ref{sec:evaluation}), and Training Data (Sec.\ref{sec:training_data}).

\subsection{Basic Focus}
\label{sec:dataset}

\begin{figure}[!t]
    \centering
    \includegraphics[width=\linewidth,scale=1.00]{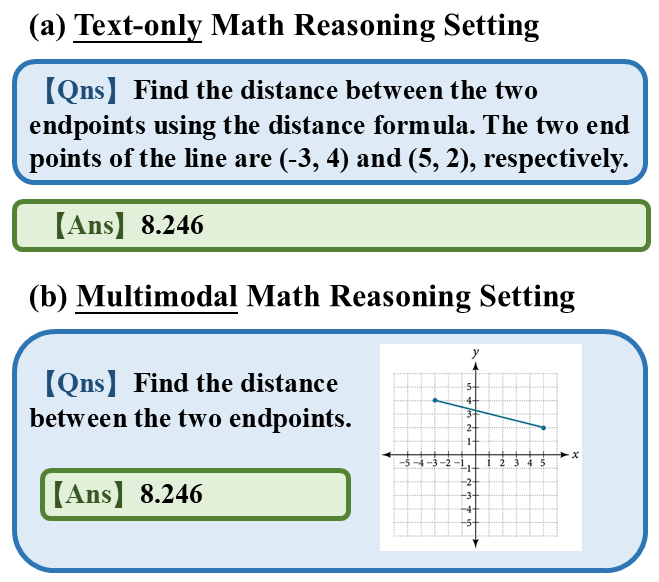}
    \caption{Typical data format of math reasoning task for text-only \& multimodal settings. Examples are derived from MathVerse \cite{zhang2025mathverse}, which assess whether and how much MLLMs can truly understand the visual diagrams for mathematical reasoning.}
    \label{fig:data_format}
    \vspace{-3mm}
\end{figure}

\textbf{Basic Format}. In a math reasoning task (taking problem-solving as a basic setting), the goal is to solve a mathematical problem given a specific format of input and output. The input consists of a statement that describes the problem to be solved. As shown in Figure \ref{fig:data_format}, this can be presented in either a textual format or a multimodal format (text accompanied by visual elements, such as figures or diagrams). The output is the predicted solution to the problem, represented as numerical or symbolic results. More cases can be seen in Appendix \ref{app:cases}.

\textbf{Language \& Size}. The majority of benchmarks are available in English, with a few exceptions like Chinese \cite{li2024cmmath} or Romanian \cite{cosma2024romath} datasets. This predominance of English datasets underscores the challenges of multilingual representation in the mathematical reasoning domain, suggesting an opportunity for future work to diversify datasets across languages, especially those in underrepresented regions. Moreover, the size of these datasets varies widely, from smaller sets (\textit{e.g.}, QRData \cite{liu2024llms} with 411 questions) to massive corpora (\textit{e.g.}, OpenMathInstruct-1 \cite{toshniwal2024openmathinstruct} with 1.8 million problem-solution pairs). Larger datasets are more likely to support robust model training and evaluation, but their size can also present challenges in terms of computational requirements and quality control.

\textbf{Source}. The sources of datasets predominantly consist of public (\textit{i.e.}, derived from public repositories or datasets) and private sources. The private datasets typically offer specialized problem types and tasks, and may present unique challenges, such as restricted access or ethical considerations. On the other hand, public datasets foster wider community collaboration, though they may suffer from limitations in diversity and task coverage. Some works have also leveraged LLMs to generate the datasets tailored to specific needs. For instance, GeomVerse constructs synthetic datasets to evaluate the multi-hop reasoning abilities required in geometric math problems \cite{kazemi2023geomverse}. 

\textbf{Educational Level}. The benchmarks span various educational levels, ranging from elementary school to university-level problems. Besides, there has also been a surge in datasets focused on competition-level problems \cite{tsoukalas2024putnambench}, offering insights into the current limitations of LLMs in comparison to the upper bound of human cognitive abilities. Future directions could involve more focused datasets targeting specific educational levels to enable models to specialize in handling particular age groups or skill sets.

\subsection{Task}
\label{sec:task}

\textbf{Model Choice}. The choice of models in these benchmarks spans open-source and closed-source models, with a growing interest in Math-LLMs. This trend indicates an increasing recognition of the need for models tailored to mathematical reasoning, which often require specialized training and handling of structured knowledge. Additionally, with the recent release of GPT-4o \cite{openai2024gpt4o} and Gemini-Pro-1.5 \cite{reid2024gemini}, which have demonstrated significant advancements in multimodal reasoning capabilities, the latest benchmarks have begun to include them in the evaluations. For example, ErrorRadar, in its initial formulation of multimodal error detection setting, incorporates these state-of-the-art MLLMs to highlight the real-world performance gap between AI systems and human-level reasoning \cite{yan2024errorradar}.

\begin{figure*}[t!]
    \centering
    \includegraphics[width=\linewidth,scale=1.00]{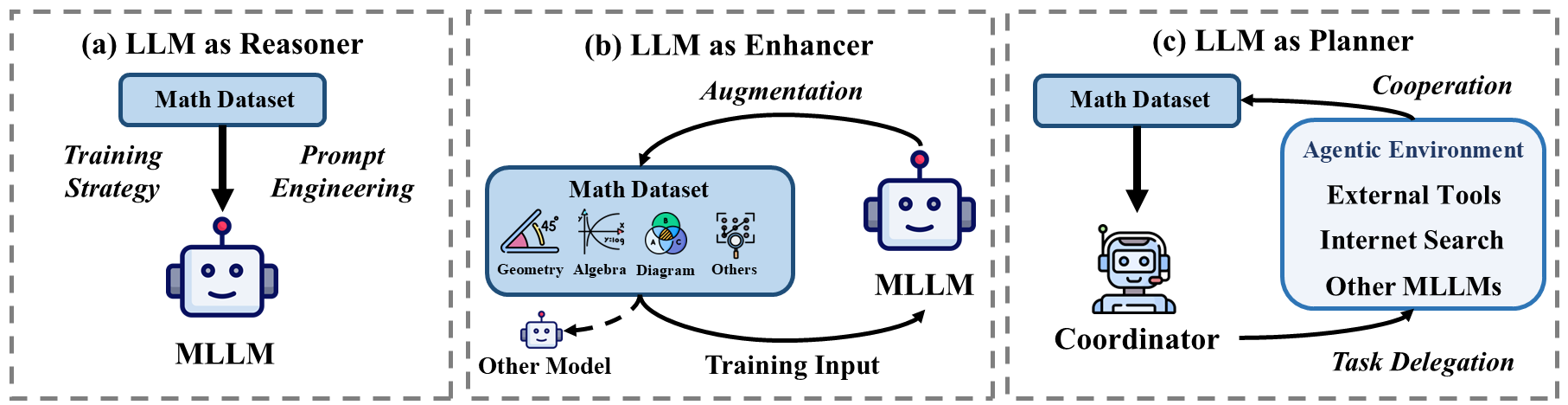}
    \caption{The illustration of the comparisons among three paradigms of (M)LLM-based mathematical reasoning.}
    \label{fig:method_comparison}
    \vspace{-4mm}
\end{figure*}

\textbf{Reasoning Task}. Problem-solving tasks typically dominate, reflecting the emphasis on students' ability to apply knowledge and reasoning skills in real-world contexts. This also serves as the core objective of current Math-LLMs. In addition, a growing proportion of error detection tasks suggests an increasing focus on helping students recognize and correct mistakes \cite{li2024evaluating,yan2024errorradar,kurtic2024mathador}. Meanwhile, proving tasks, often associated with higher-order thinking, highlight a shift towards cultivating logical reasoning and systematic problem-solving abilities \cite{tsoukalas2024putnambench}. Moreover, a smaller portion of work has addressed tasks that align with real-world educational needs but lack systematic formulation. For instance, \citet{li2024evaluating} further introduces error correction (which goes beyond simple error detection); \citet{didolkar2024metacognitive} explores automated skill discovery for problem-solving; and MathChat \cite{liang2024mathchat} focuses on reasoning in multi-turn settings (such as follow-up QA and problem generation). Given the higher demands on reasoning capabilities in multimodal settings, many studies have also evaluated the aforementioned reasoning tasks in image-text problem settings. These efforts aim to provide the LLM community with more diverse, real-world task scenarios, catering to the needs of multimodal learning environments.

\subsection{Evaluation}
\label{sec:evaluation}

\textbf{Discriminative Evaluation} is a common approach, focusing on the ability of M(LLM)s to correctly classify or choose the correct answer \cite{hendrycks2021measuring, mishra2022lila,li2024mugglemath}. Based on specific motivations, some works also build their metrics upon accuracy for further expansion. For example, GSM-PLUS, a new adversarial benchmark for evaluating the robustness of LLMs in mathematical reasoning, develops performance drop rate (PDR) to measure the relative decline in performance on question variations compared to the original questions \cite{li2024gsm}. ErrorRadar uses error step accuracy and error category accuracy together to evaluate the multimodal error detection of MLLMs \cite{yan2024errorradar}.

\textbf{Generative Evaluation}, on the other hand, measures a M(LLM)’s ability to produce detailed explanations or solve problems from scratch. This evaluation type is gaining traction, particularly for complex mathematical tasks where step-by-step solutions are required. For instance, MathVerse, which modifies problems with varying degrees of information content in multi-modality, employs GPT-4 to score each key step in the reasoning process generated by MLLMs \cite{zhang2025mathverse}. CHAMP proposes a solution evaluation pipeline where GPT-4 is utilized as a grader for the answer summary, given the ground truth answer \cite{mao2024champ}.

Due to page limit, more details of both types of evaluation metrics can be seen in Appendix \ref{app:metrics}.

\subsection{Training Data}
\label{sec:training_data}
The training of MLLMs for mathematical reasoning relies on a carefully orchestrated integration of \textit{instruction design}, \textit{data scale}, and \textit{task diversity} to ensure robust and generalizable performance. Central to this process is the \textbf{design of instruction sets}, which are structured to bridge symbolic, textual, and visual reasoning \cite{toshniwal2024openmathinstruct,toshniwal2024openmathinstruct2}. These instructions progressively escalate in complexity, starting from foundational arithmetic to advanced domains like calculus and linear algebra, ensuring models build skills incrementally. Each problem can be accompanied by explicit step-by-step explanations, enabling models to learn logical sequencing and self-correction \cite{zhang2024mavis,tang2024mathscale,liang2024mathchat}. 

The \textbf{scale of pre-training data} also plays an equally critical role. Models are exposed to terabytes of data sourced from textbooks, research papers (\textit{e.g.,} arXiv), online educational platforms (\textit{e.g.,} Khan Academy), and synthetically generated problems. A significant portion (10–30\%) of the pretraining corpus is dedicated to mathematical content, with specialized datasets ensuring coverage of niche topics. While scaling to trillion-token corpora enhances robustness, rigorous filtering mechanisms, such as self-supervised quality checks, are applied to eliminate noise, including incorrect solutions or irrelevant content \cite{shao2024deepseekmath,qwen2math,yue2024mammoth2}.

Finally, the \textbf{variety of mathematical tasks} ensures models adapt to diverse challenges. Training spans core domains like algebra and geometry, as well as cross-disciplinary applications (\textit{e.g.,} physics-based calculus problems). Tasks are presented in multiple formats: closed-ended questions (\textit{e.g.,} solving equations), open-ended prompts (\textit{e.g.,} deriving proofs), and error-analysis exercises that require identifying and correcting flawed reasoning \cite{lu2022survey,yan2025position}.

For example, G-LLaVA \cite{gao2023gllava} focuses on solving geometry problems by extracting visual features from geometric figures and jointly modeling them with text descriptions, allowing the model to understand key elements (\textit{e.g.,} points, lines, angles) in geometric figures and their relationship with text descriptions. MAVIS \cite{zhang2024mavis} features an automatic data generation engine that can quickly generate large-scale, high-quality multimodal mathematical datasets, addressing the problem of data scarcity. It also uses instruction fine-tuning to teach the model how to decompose complex mathematical problems and generate reasonable reasoning steps (\textit{esp.,} MAVIS-Instruct includes 834k visual math problems with CoT rationales). Math-LLaVA \citep{shi2024math} uses the MathV360K multimodal dataset (360k instances), which covers multiple mathematical domains to gradually improve the model’s mathematical reasoning ability through bootstrapping and further optimize the model using generated data.

\section{Methodology Perspective}
\label{sec:method}

\subsection{Overview \& Findings}
MLLMs have been leveraged in various ways to tackle the broad spectrum of mathematical reasoning tasks. Based on our comprehensive review of recent methodologies (summarized in Table \ref{tab:method} from Appendix \ref{app:methods}), we classify the works into three distinct paradigms: LLM as Reasoner (Sec.\ref{sec:llm_reasoner}), LLM as Enhancer (Sec.\ref{sec:llm_enhancer}), and LLM as Planner (Sec.\ref{sec:llm_planner}), and finally provide a in-depth comparison of technical distinctions (Sec.\ref{sec:paradigm_comparison}).

\textbf{Findings}. First, single-modality settings dominate the current landscape of method-oriented research, with the majority focusing solely on algebraic tasks. However, since 2024, multimodal approaches have been increasingly incorporated, expanding the scope of mathematical reasoning to include geometry, diagrams, and even broader mathematical concepts. This shift signals a growing interest in enhancing model robustness through multimodal learning, which can address the diverse nature of mathematical problems. Second, regarding the evaluated tasks, problem-solving and proving are gaining prominence, while some research also focuses on error detection or others (\textit{e.g.}, RefAug includes error correction and follow-up QA as evaluation tasks \citep{zhang2024learn}). Finally, in terms of the role of LLMs, Reasoner is the most common role, followed by Enhancer, while Planner remains less explored but holds promise due to recent advancements in multi-agent intelligence.

\subsection{LLM as Reasoner}
\label{sec:llm_reasoner}
\textbf{Definition}. In the \textit{Reasoner} paradigm, M(LLM)s harness their inherent reasoning capabilities to solve mathematical problems, as shown in Figure \ref{fig:method_comparison} (a). This can either involve fine-tuning existing LLMs on task-specific datasets or utilizing zero-shot or few-shot learning strategies. These models utilize advanced semantic understanding and reasoning techniques, such as symbolic manipulation, logical deduction, and multi-step reasoning.

\textbf{Examples}. \citet{deng2023towards} develops a unified framework for answer calibration that integrates step-level and path-level strategies on multi-step reasoning of LLMs. MATH-SHEPHERD serves as a process-oriented math verifier, which assigns a reward score to each step of the LLM's outputs on math questions \citep{wang2024math}. As for multimodal approaches, Math-PUMA introduces progressive upward multimodal alignment strategy for reasoning-enhanced training \citep{zhuang2024math}; Math-LLaVA, a LLaVA-1.5-based model, directly bootstraps mathematical reasoning via fine-tuned on 360K high-quality math QA pairs, which can ensure the depth and breadth of multimodal mathematical problems \cite{shi2024math}; STIC develops a two-stage self-training pipeline (consisting of Image Comprehension Self-Training phase \& Description-Infused Fine-Tuning phase) for enhancing visual comprehension \citep{deng2024enhancing}; VCAR emphasizes on the visual-centric supervision, thus proposing a similar two-step training pipleine which handles the visual description generation task first, followed by mathematical rationale generation task \citep{jia2024describe}.

\textbf{Summary \& Outlook}. This paradigm has shown significant promise, particularly in solving problems requiring multiple steps of reasoning. However, despite improvements, issues with robustness remain, particularly with zero-shot reasoning tasks. Future work should focus on combining reasoning with structured knowledge retrieval systems and enhancing models' ability to reason effectively across diverse domains, especially in multimodal contexts \cite{fan2024survey,pan2023large}.

\subsection{LLM as Enhancer}
\label{sec:llm_enhancer}
\textbf{Definition}. In the \textit{Enhancer} paradigm, M(LLM)s are primarily used to augment data, thereby enabling improvements in mathematical reasoning, as illustrated in Figure \ref{fig:method_comparison} (b). This can be achieved by synthesizing new training data, refining existing datasets, or introducing new variations that target specific problem-solving abilities \cite{li2022data}. Data augmentation can include paraphrasing mathematical problems, adding noise to mathematical expressions, or generating problem variants for underrepresented cases.

\textbf{Examples}. A typical example of a single-modality enhancement approach is Masked Thought, which introduces perturbations to the input and randomly masks tokens within the chain of thought during training \citep{chen2024masked}. MathGenie, which aims to generate diverse and reliable math problems and solution from a small-scale dataset, leverages a solution augmentation model to iteratively create new solutions from existing ones \cite{lu2024mathgenie}. For multimodal methods, AlphaGeometry proves most olympiad-level mathematical theorems, via trained from scratch on large-scale synthetic data guiding the symbolic deuction \cite{trinh2024solving}; LogicSolver introduces interpretable formula-based tree-structure for each solution equation \cite{yang2022logicsolver}; InfiMM-Math achieves the exceptional performance as it is trained on a large-scale multimodal interleaved math dataset developed and validated by LLMs such as LLaMA3-70B-Instruct \cite{han2024infimm}; DFE-GPS constructs its synthetic training set, which integrates visual features and geometric formal language \citep{zhang2024diagram}.

\textbf{Summary \& Outlook}. This paradigm offers substantial performance improvements by enriching the training set. However, challenges remain in ensuring the diversity and relevance of the generated data. Moreover, while text-based augmentation methods have proven effective, the potential for multimodal augmentation is still underexplored. Future research should focus on advancing multimodal data augmentation techniques, especially for tasks that require interaction between visual and textual modalities \cite{xiao2023multimodal}.

\subsection{LLM as Planner}
\label{sec:llm_planner}
\textbf{Definition}. In the \textit{Planner} paradigm, M(LLM)s are treated as coordinators that guide the solution of complex mathematical problems by delegating tasks to other models or tools, as illustrated in Figure \ref{fig:method_comparison} (c). This includes scenarios where multiple agents or models collaborate to achieve a single objective, thereby enhancing the performance of mathematical problem-solving through cooperative interactions. These models often work in environments with multiple steps or require iterative refinement of solutions.

\begin{table*}[!t] 
    \centering
    \resizebox{2\columnwidth}{!}{
    \begin{tabular}{lccc}  
    \toprule
    \textbf{Aspect}  & \textbf{LLM as Reasoner} &  \textbf{LLM as Enhancer} & \textbf{LLM as Planner} \\
    \midrule  
    \textbf{Data Interaction Patterns} &  &  &  \\
    \midrule
    \textit{Input-Output Relation} & \makecell{End-to-end mapping\\ (Problem → Answer)} & \makecell{Data augmentation pipeline\\ (Raw data → Enhanced data)} & \makecell{Dynamic workflow planning\\ (Problem → Plan → Subtasks)}  \\
    \textit{External Dependencies} & \makecell{Low \\(Self-contained reasoning)} & \makecell{Medium \\(Data distribution dependent)} & \makecell{High \\(Requires toolchain integration)}\\
    \midrule
    \textbf{Pros \& Cons} &  &  &  \\
    \midrule
    \textit{Advantages} & \makecell{Transparent reasoning \& \\Strong interpretability} & \makecell{Improves generalization \& \\Handles data scarcity} & \makecell{Breaks capability boundaries \& \\Enables complex task solving} \\
    \textit{Limitations} & Error-prone in complex reasoning & May introduce semantic biases & High system complexity \& Increased latency  \\
    \bottomrule    
    \end{tabular}}
    \caption{Comparisons among the three methodology paradigms.} 
    \label{tab:paradigm_comparison}
    \vspace{-4mm}
\end{table*}

\textbf{Examples}. A notable tool-integrated agent is ToRA, which plans the sequential use of natural language rationale and program-based tools synergistically to solve mathematical problems in an optimal manner \cite{gou2023tora}. Additionally, COPRA simulates a single agent-like reasoning mechanism where GPT-4 proposes tactic applications within a stateful backtracking search, leveraging feedback from the proof environment \citep{thakur2024context}. This can also extend to multimodal scenarios, as seen in Chameleon, which serves as an AI system that augments MLLMs with plug-and-play modules for compositional reasoning, leveraging an LLM-based planner to assemble tools for complex tasks \citep{lu2024chameleon}. Furthermore, Visual Sketchpad presents the concept of sketching as a ubiquitous tool used by humans for communication, ideation, and problem-solving. Hence, MLLMs can enable external tools (\textit{e.g.}, matplotlib) to generate intermediate sketches to aid in reasoning, which includes an iterative interaction process with an environment \citep{hu2024visual}. Although there has been much work on Compositional Visual Reasoning in the past \cite{gupta2023visual,suris2023vipergpt,yao2022react}, Visual Sketchpad is the first work that integrates the planning capabilities of MLLMs with the real gap of mathematical reasoning settings (\textit{i.e.}, sketch-based reasoning involving visuo-spatial concepts).

\textbf{Summary \& Outlook}. While the Planner paradigm introduces significant improvements, particularly for complex tasks that require multi-agent collaboration, it remains a relatively under-explored area \cite{xi2023rise,guo2024large}. There is potential for further improvement in task decomposition, agent cooperation strategies, and integration of diverse computational tools. Future work will likely focus on refining these planning strategies, especially for multimodal systems that can jointly leverage visual and textual knowledge to solve more intricate problems \cite{yan2025mathagent,durante2024agent,li2023multimodal}.

\subsection{Paradigm Comparison}
\label{sec:paradigm_comparison}

As summarized in Table \ref{tab:paradigm_comparison}, we list the differences between the three paradigms to provide the community with a more comprehensive understanding of the latest technical distinctions. These three paradigms show a progressive development logic: Reasoner focuses on intrinsic model capabilities, Enhancer targets data optimization, and Planner moves towards system-level intelligent collaboration. In practice, we also anticipate adopting a hybrid approach (\textit{e.g., }using Enhancer to generate augmented data to train Reasoner, then coordinating multiple Reasoner modules via Planner to solve complex problems). This layered architecture may become the core design paradigm for future multimodal mathematical reasoning systems.

\section{Challenges}
\label{sec:challenge}
In the realm of MLLMs for mathematical reasoning, the following key challenges persist that hinder their full potential. Addressing these challenges is essential for advancing MLLMs toward more robust and flexible systems that can better support mathematical reasoning in real-world settings.

\ding{182} \textbf{Lack of High-Quality, Diverse, and Large-Scale Multimodal Datasets}. As discussed in Section \ref{sec:training_data}, current multimodal mathematical reasoning datasets face tripartite limitations in quality (\textit{e.g.,} misaligned text-image pairs), scale (insufficient advanced topic coverage), and task diversity (overemphasis on problem-solving versus error diagnosis or theorem proving). For instance, most datasets focus on question answering but lack annotations for error tracing steps or formal proof generation, while synthetic datasets often exhibit domain bias \cite{wang2024measuring,lu2023mathvista}. Three concrete solutions emerge: i) Develop hybrid dataset construction pipelines combining expert-curated problems with AI-augmented task variations; ii) Implement cross-task knowledge distillation, where models trained on proof generation guide error diagnosis through attention pattern transfer; iii) Leverage automated frameworks quality-controlled multimodal expansion to systematically generate diverse task formats (\textit{e.g.,} converting proof exercises into visual dialogues). More discussion on data bottlenecks in Appendix \ref{app:data_bottleneck}. 

\ding{183} \textbf{Insufficient Visual Reasoning}. Many math problems require extracting and reasoning over visual content, such as charts, tables, or geometric diagrams. Current models struggle with intricate visual details, such as interpreting three-dimensional geometry or analyzing irregularly structured tables \cite{zhang2025mathverse}. Hence, it may be beneficial to introduce enhanced visual feature extraction modules and integrate scene graph representations for better reasoning over complex visual elements \cite{ibrahim2024survey,guo2024knowledgenavigator}.

\ding{184} \textbf{Reasoning Beyond Text and Vision}. While the current research focus on the combination of text and vision, mathematical reasoning in real-world applications often extends beyond these two modalities. For instance, audio explanations, interactive problem-solving environments, or dynamic simulations might play a role in some tasks. Current models are not well-equipped to handle such diverse inputs \cite{abrahamson2020future,jusslin2022embodied}. To address this, datasets should be expanded to include more diverse modalities, such as audio, video, and interactive tools. MLLMs should also be designed with flexible architectures capable of processing and reasoning over multiple types of inputs, allowing for a richer representation of mathematical problems \cite{dasgupta2023collaborating}.

\ding{185} \textbf{Limited Domain Generalization}. Mathematical reasoning spans many domains, such as algebra, geometry, diagram and commonsense, each with its own specific requirements for problem-solving \cite{liu2023mathematical,lu2022survey}. Math-LLMs that perform well in one domain often fail to generalize across others, which can limit their utility. By pretraining and fine-tuning Math-LLMs on a wide array of problem types, models may handle cross-domain tasks more effectively, improving their ability to generalize across different mathematical topics and problem-solving strategies. We extend more discussion on limited domain generalization in multimodal contexts in Appendix \ref{app:limited_domain_generalization}.

\ding{186} \textbf{Error Feedback Limitations}. Mathematical reasoning involves various types of errors, such as calculation mistakes, logical inconsistencies, and misinterpretations of the problem. Currently, MLLMs lack mechanisms to detect, categorize, and correct these errors effectively, which can result in compounding mistakes throughout the reasoning process \cite{yan2024errorradar,li2024evaluating}. A potential solution is to integrate error detection and classification modules that can identify errors at each step of the reasoning process. Besides, multi-agent collaboration mechanism could be introduced, via involving multiple agents collaborating by exchanging feedback and collectively refining the reasoning process \cite{xu2024ai}. We extend more discussion on error feedback limitation in multimodal contexts in Appendix \ref{app:error_feedback_limitation}.

\ding{187} \textbf{Integration with Real-world Educational Needs}. Existing benchmarks and models often overlook real-world educational contexts, such as how students use draft work, like handwritten notes or diagrams, to solve problems \cite{xu2024foundation,wang2024large}. These real-world elements are crucial for understanding how humans approach mathematical reasoning \cite{mouchere2011crohme2011, gervais2024mathwriting}. By incorporating draft notes, handwritten calculations, and dynamic problem-solving workflows into the training data, MLLMs can be tailored to provide more accurate and contextually relevant feedback for students.

\ding{188} \textbf{Test-Time Scaling Technique in Multimodal Context}. While foundation models increasingly adopt test-time scaling techniques (\textit{e.g.,} dynamic architecture adaptation), their integration with multimodal mathematical reasoning remains underexplored and suboptimal. For example, current implementations like o1 \cite{jaech2024openai} or DeepSeek-R1 \cite{guo2025deepseekr1} struggle to dynamically allocate computational resources based on math problem complexity across modalities, such as deciding when to prioritize symbolic computation over visual parsing for optimization problems. Future work should focus on two directions: i) Develop modality-aware scaling controllers that jointly consider problem type, visual complexity, and required mathematical operations to optimize dynamic architecture decisions; ii) Create lightweight meta-optimization layers that can adjust model capacity allocation (\textit{e.g.,} expert selection in MoE systems) through real-time analysis of multimodal problem-solving workflows \cite{xu2025towards,besta2025reasoning}. Such advancements could enable more efficient trade-offs between accuracy and computational cost in deployed systems. We also discuss how test-time scaling techniques can tackle the other challenges in Appendix \ref{app:test_time_scaling}.

\vspace{-1mm}
\section{Conclusion}
\label{sec:conclusion}
In this survey, we have provided a comprehensive overview of the progress and challenges in mathematical reasoning within the context of MLLMs. We highlighted the significant advances in the development of Math-LLMs and the growing importance of multimodal integration for solving complex reasoning tasks. We identified five key challenges that are crucial for the continued development of AGI systems capable of performing sophisticated mathematical reasoning tasks. As research continues to advance, it is essential to focus on these challenges to unlock the full potential of LLMs in multimodal settings. We hope this survey provides insights to guide future LLM research, ultimately leading to more effective and human-like mathematical reasoning capabilities in AI systems.

\clearpage
\section*{Limitations}
Despite our best efforts to ensure comprehensive coverage of the published works, it is possible that some relevant studies were overlooked. Additionally, human errors could have occurred during the categorization or referencing of papers in the survey. To minimize such errors, we made a concerted effort to gather studies from multiple sources and performed a multiple-round checking process. While minor inconsistencies or omissions may still exist, we believe this survey represents the most comprehensive review of MLLM-based mathematical reasoning to date, effectively capturing key research trends and highlighting ongoing challenges.

\section*{Acknowledgements}
This work was supported by Guangdong Provincial Department of Education Project (Grant No.2024KQNCX028); Scientific Research Projects for the Higher-educational Institutions (Grant No.2024312096), Education Bureau of Guangzhou Municipality; Guangzhou-HKUST(GZ) Joint Funding Program (Grant No.2025A03J3957), Education Bureau of Guangzhou Municipality.

\bibliography{survey}

\clearpage
\appendix

\section{Details of Math-LLMs' Progress}
\label{app:math_llm}

The rapid development of general-purpose LLMs has made significant advancements in natural language processing tasks. However, the development of domain-specific models remains a core requirement, as they are better equipped to handle specialized tasks that general models may not address effectively \cite{ge2023openagi,shen2024tag,huo2024mmneuron,huo2025mmunlearner,chen2025safeeraser}. This is particularly true in fields such as healthcare \cite{liu2023qilin,nazi2024large}, law \cite{cui2023chatlaw,zhou2024lawgpt,wang2023exploring}, finance \cite{wu2023bloomberggpt,yang2023fingpt,zhang2023xuanyuan}, education \cite{ye2025position,su2025essayjudge,gupta2025beyond}, and urban science \cite{yan2024urbanclip,zou2025deep,yan2024georeasoner}, where domain-specific knowledge is critical for high accuracy and performance.

In the case of mathematical reasoning, general models may struggle with tasks that require a deep understanding of complex mathematical concepts, structures, and problem-solving steps \cite{yan2025position}. Therefore, the development of math-specific LLMs is of paramount importance, as these models are designed to enhance performance in mathematical reasoning, theorem proving, equation solving, and other math-intensive tasks.

Therefore, Table \ref{tab:math-llm} provides a detailed overview of various math-specific LLMs (\textit{i.e.}, Math-(LLMs), sorted by their release date. It includes information about the organization behind each model, the release date, publication details, language(s) supported, parameter size, evaluation benchmarks, and whether the model is open source.

Key findings are summarized as follows:
\begin{enumerate}
    \item \textbf{Release Trends:} The models started emerging in 2020, with a significant increase in the number of releases from 2022 onward, indicating a growing interest in developing math-specific LLMs.
    \item \textbf{Parameter Sizes:} There is a noticeable trend towards larger parameter sizes, with some models offering up to 130B parameters, reflecting the increasing computational capacity for handling complex mathematical tasks.
    \item \textbf{Evaluation Benchmarks:} Many models are evaluated on popular benchmarks like GSM8K, MATH, and MMLU, highlighting the focus on improving performance across well-established mathematical reasoning datasets.
    \item \textbf{Multilingual Support:} While most models are focused on English, a few (\textit{e.g.}, MathGPT \& Math-LLM) also support Chinese, showing a trend towards multilingual capabilities.
    \item \textbf{Open Source:} A significant number of models are open-source, allowing broader access and fostering further research and development in the field.
\end{enumerate}

In summary, the table reflects the rapid development of specialized Math-LLMs, with an increasing trend towards larger models, comprehensive evaluation benchmarks, and support for multilingual applications.

\begin{table*}[ht!]
\tiny
\centering
\resizebox{\linewidth}{!}{
    \begin{tabular}{lcccccccc} 
    \toprule
    \textbf{Benchmarks} & \textbf{Venue} & \textbf{Language} & \textbf{Size} & \textbf{Source} & \textbf{Level(s)} & \textbf{Evaluation} & \textbf{Model(s)} & \textbf{Task(s)}  \\
    \midrule
    DynaMath \citep{zou2024dynamath} \FiveStarConvex & ICLR'25 & English & 5,010 & \encircle[fill=DarkBlue, text=white]{S} \hspace{2pt}\encircle[fill=DarkBlue, text=white]{P} \hspace{2pt}\encircle[fill=DarkBlue, text=white]{G} &  \encircle[fill=MyGreenv2, text=white]{E} \hspace{2pt}\encircle[fill=myBlue, text=white]{M} \hspace{2pt}\encircle[fill=myYellowv2, text=white]{H} \hspace{2pt}\encircle[fill=myOrangev2, text=white]{U} & Both & Closed/Open &  \encircle[fill=DarkRed, text=white]{S} \\

    MathCheck \citep{zhou2024your} \FiveStarConvex & ICLR'25 & English/Chinese & 4,536 & \encircle[fill=DarkBlue, text=white]{P} & \encircle[fill=MyGreenv2, text=white]{E} \hspace{2pt}\encircle[fill=myBlue, text=white]{M} \hspace{2pt}\encircle[fill=myYellowv2, text=white]{H} \hspace{2pt}\encircle[fill=myOrangev2, text=white]{U} & Discriminative & Closed/Open/Math & \encircle[fill=DarkRed, text=white]{S}  \\

    GSM-Symbolic \citep{mirzadeh2024gsm} & ICLR'25 & English & 5,000 & \encircle[fill=DarkBlue, text=white]{P} & \encircle[fill=MyGreenv2, text=white]{E} & Discriminative & Closed/Open & \encircle[fill=DarkRed, text=white]{S} \\

    Omni-MATH \citep{gao2024omni} & ICLR'25 & English & 4,428 & \encircle[fill=DarkBlue, text=white]{S} & \encircle[fill=myViolet, text=white]{C} & Discriminative & Closed/Open/Math & \encircle[fill=DarkRed, text=white]{S} \\

    HARDMath \citep{fan2024hardmath} & ICLR'25 & English & 1,466 & \encircle[fill=DarkBlue, text=white]{G} & \encircle[fill=myOrangev2, text=white]{U} & Both & Closed/Open & \encircle[fill=DarkRed, text=white]{S} \\

    OpenMathInstruct-2 \citep{toshniwal2024openmathinstruct2} & ICLR'25 & English & 14,000,000 & \encircle[fill=DarkBlue, text=white]{P} \hspace{2pt}\encircle[fill=DarkBlue, text=white]{G} & \encircle[fill=MyGreenv2, text=white]{E}  \hspace{2pt}\encircle[fill=myYellowv2, text=white]{H} \hspace{2pt}\encircle[fill=myOrangev2, text=white{U}
    \hspace{2pt}\encircle[fill=myViolet, text=white]{C} & Discriminative & Open/Math & \encircle[fill=DarkRed, text=white]{S} \\

    UGMathBench \citep{xu2025ugmathbench} & ICLR'25 & English & 5,062 & \encircle[fill=DarkBlue, text=white]{S} & \encircle[fill=myOrangev2, text=white]{U} & Discriminative & Closed/Open/Math & \encircle[fill=DarkRed, text=white]{S} \\

    ErrorRador \citep{yan2024errorradar} \FiveStarConvex & ICLR Workshop'25 & English & 2,500 & \encircle[fill=DarkBlue, text=white]{S} & \encircle[fill=MyGreenv2, text=white]{E} \hspace{2pt}\encircle[fill=myBlue, text=white]{M} \hspace{2pt}\encircle[fill=myYellowv2, text=white]{H} & Discriminative & Closed/Open & \encircle[fill=DarkRed, text=white]{D}  \\

    $\mathsf{M}^3\mathsf{CoT}_{\text{math}}$  \citep{chen2024m} \FiveStarConvex & ACL'24 & English & 1,166 & \encircle[fill=DarkBlue, text=white]{P} \hspace{2pt}\encircle[fill=DarkBlue, text=white]{G} & \encircle[fill=myViolet, text=white]{C} & Discriminative & Closed/Open & \encircle[fill=DarkRed, text=white]{S}   \\ 
    
    GSM-Plus \citep{li2024gsm} & ACL'24 & English & 10,552 & \encircle[fill=DarkBlue, text=white]{P} & \encircle[fill=MyGreenv2, text=white]{E} \hspace{2pt}\encircle[fill=myBlue, text=white]{M} \hspace{2pt}\encircle[fill=myYellowv2, text=white]{H} \hspace{2pt}\encircle[fill=myOrangev2, text=white]{U} & Generative & Closed/Open/Math & \encircle[fill=DarkRed, text=white]{S}   \\ 
    
    MuggleMath \citep{li2024mugglemath} & ACL'24 & English & 37,365 & \encircle[fill=DarkBlue, text=white]{P} & \encircle[fill=MyGreenv2, text=white]{E}
    \hspace{2pt}\encircle[fill=myYellowv2, text=white]{H} & Discriminative & Open & \encircle[fill=DarkRed, text=white]{S}   \\
    
    Olympiadbench \citep{he2024olympiadbench} \FiveStarConvex & ACL'24 & English/Chinese & 8,476 & \encircle[fill=DarkBlue, text=white]{S} & \encircle[fill=myYellowv2, text=white]{H} \hspace{2pt}\encircle[fill=myViolet, text=white]{C} & Generative & Closed/Open/Math & \encircle[fill=DarkRed, text=white]{S}   \\
    
    MathBench \citep{liu2024mathbench} & ACL Findings'24 & English/Chinese & 3,709 & \encircle[fill=DarkBlue, text=white]{P} \hspace{2pt}\encircle[fill=DarkBlue, text=white]{S} & \encircle[fill=MyGreenv2, text=white]{E} \hspace{2pt}\encircle[fill=myBlue, text=white]{M} \hspace{2pt}\encircle[fill=myYellowv2, text=white]{H} \hspace{2pt}\encircle[fill=myOrangev2, text=white]{U} & Generative & Closed/Open/Math & \encircle[fill=DarkRed, text=white]{S}   \\
     
    GeoEval \citep{zhang2024geoeval} \FiveStarConvex & ACL Findings'24 & English & 5,050  & \encircle[fill=DarkBlue, text=white]{P} \hspace{2pt}\encircle[fill=DarkBlue, text=white]{G} & \encircle[fill=MyGreenv2, text=white]{E} \hspace{2pt}\encircle[fill=myBlue, text=white]{M} \hspace{2pt}\encircle[fill=myYellowv2, text=white]{H} & Discriminative & Closed/Open/Math & \encircle[fill=DarkRed, text=white]{S}   \\
     
    QRData \citep{liu2024llms} & ACL Findings'24 & English & 411 & \encircle[fill=DarkBlue, text=white]{S} & \encircle[fill=myOrangev2, text=white]{U} & Discriminative & Closed/Open/Math & \encircle[fill=DarkRed, text=white]{S}\\
     
    EIC-Math \citep{li2024evaluating} & ACL Findings'24 & English & 1,800 & \encircle[fill=DarkBlue, text=white]{P} & \encircle[fill=MyGreenv2, text=white]{E} \hspace{2pt}\encircle[fill=myBlue, text=white]{M} \hspace{2pt}\encircle[fill=myYellowv2, text=white]{H} & Discriminative & Closed/Open & \encircle[fill=DarkRed, text=white]{D} \hspace{2pt}\encircle[fill=DarkRed, text=white]{O}     \\

    \citet{srivastava2024evaluating} & ACL Findings'24 & English & - & \encircle[fill=DarkBlue, text=white]{P} & \encircle[fill=MyRedv2, text=white]{H} & Discriminative & Closed/Open & \encircle[fill=DarkRed, text=white]{S}   \\

    CHAMP \citep{mao2024champ} & ACL Findings'24 & English & 270 & \encircle[fill=DarkBlue, text=white]{S} & \encircle[fill=myYellowv2, text=white]{H} & Generative & Closed/Open & \encircle[fill=DarkRed, text=white]{S}  \\
    
    IMO-AG-30 \citep{trinh2024solving} & Nature'24 & English & 30 & \encircle[fill=DarkBlue, text=white]{S} & \encircle[fill=myViolet, text=white]{C} & Discriminative & Closed & \encircle[fill=DarkRed, text=white]{P}  \\

    PutnamBench \citep{tsoukalas2024putnambench} & NeurIPS'24 & English & 1,697 & \encircle[fill=DarkBlue, text=white]{S} & \encircle[fill=myViolet, text=white]{C} & Generative & Closed & \encircle[fill=DarkRed, text=white]{S} \hspace{2pt}\encircle[fill=DarkRed, text=white]{P} \\

    MATH-Vision \citep{wang2024measuring} \FiveStarConvex & NeurIPS'24 & English & 3,040 & \encircle[fill=DarkBlue, text=white]{S} & \encircle[fill=MyGreenv2, text=white]{E} \hspace{2pt}\encircle[fill=myBlue, text=white]{M} \hspace{2pt}\encircle[fill=myYellowv2, text=white]{H} \hspace{2pt}\encircle[fill=myOrangev2, text=white]{U} & Discriminative & Closed/Open & \encircle[fill=DarkRed, text=white]{S} \\

    CARP \citep{zhang2024evaluating} & NeurIPS'24 & Chinese & 4,886 & \encircle[fill=DarkBlue, text=white]{S} & \encircle[fill=myViolet, text=white]{C} & Discriminative & Closed & \encircle[fill=DarkRed, text=white]{S} \\

    SMART-840 \citep{cherian2024evaluating} \FiveStarConvex & NeurIPS'24 & English & 840 & \encircle[fill=DarkBlue, text=white]{S} & \encircle[fill=MyGreenv2, text=white]{E} \hspace{2pt}\encircle[fill=myBlue, text=white]{M} \hspace{2pt}\encircle[fill=myYellowv2, text=white]{H} & Discriminative & Closed/Open & \encircle[fill=DarkRed, text=white]{S} \\

    OpenMathInstruct-1 \citep{toshniwal2024openmathinstruct} & NeurIPS'24 & English & 1,800,000 & \encircle[fill=DarkBlue, text=white]{P} & \encircle[fill=MyGreenv2, text=white]{E} \hspace{2pt}\encircle[fill=myBlue, text=white]{M} \hspace{2pt}\encircle[fill=myYellowv2, text=white]{H} \hspace{2pt}\encircle[fill=myViolet, text=white]{C} & Generative & Closed/Open/Math & \encircle[fill=DarkRed, text=white]{S} \\

    \citet{didolkar2024metacognitive} & NeurIPS'24 & English & 8,600 & \encircle[fill=DarkBlue, text=white]{P} & \encircle[fill=MyGreenv2, text=white]{E} & Discriminative & Closed & \encircle[fill=DarkRed, text=white]{S} \hspace{2pt}\encircle[fill=DarkRed, text=white]{O} \\

    Putnam-AXIOM \citep{gulati2024putnam} & NeurIPS Workshop'24 & English & 236 & \encircle[fill=DarkRed, text=white]{S} & \encircle[fill=myViolet, text=white]{C} & Discriminative & Closed/Open/Math & \encircle[fill=DarkRed, text=white]{S} \\

    Scibench \citep{wang2023scibench} \FiveStarConvex & ICML'24 & English & 869 & \encircle[fill=DarkBlue, text=white]{S} & \encircle[fill=myOrangev2, text=white]{U} & Discriminative & Closed/Open & \encircle[fill=DarkRed, text=white]{S} \\

    GeomVerse \citep{kazemi2023geomverse} \FiveStarConvex & ICML Workshop'24 & English & 1,000 & \encircle[fill=DarkBlue, text=white]{G} & \encircle[fill=myOrangev2, text=white]{U} & Discriminative & Closed & \encircle[fill=DarkRed, text=white]{S}  \\

    MathVista \citep{lu2023mathvista} \FiveStarConvex & ICLR'24 & English & 6,141 & \encircle[fill=DarkBlue, text=white]{S} \hspace{2pt}\encircle[fill=DarkBlue, text=white]{P} & \encircle[fill=MyGreenv2, text=white]{E} \hspace{2pt}\encircle[fill=myBlue, text=white]{M} \hspace{2pt}\encircle[fill=myYellowv2, text=white]{H} \hspace{2pt}\encircle[fill=myOrangev2, text=white]{U} & Discriminative & Closed/Open & \encircle[fill=DarkRed, text=white]{S}  \\

    $\text{MMMU}_{\text{math}}$ \citep{yue2023mmmu} \FiveStarConvex & CVPR'24 & English & 540 & \encircle[fill=DarkBlue, text=white]{S} & \encircle[fill=myOrangev2, text=white]{U} & Discriminative & Closed/Open & \encircle[fill=DarkRed, text=white]{S} \\

    MathVerse \citep{zhang2025mathverse} \FiveStarConvex & ECCV'24 & English & 2,612 & \encircle[fill=DarkBlue, text=white]{S} \hspace{2pt}\encircle[fill=DarkBlue, text=white]{P} & \encircle[fill=myYellowv2, text=white]{H} & Generative & Closed/Open & \encircle[fill=DarkRed, text=white]{S} \\

    Mathador-LM \citep{kurtic2024mathador} & EMNLP'24 & English & - & \encircle[fill=DarkBlue, text=white]{G} & \encircle[fill=MyGreenv2, text=white]{E} & Both & Closed/Open & \encircle[fill=DarkRed, text=white]{S} \hspace{2pt}\encircle[fill=DarkRed, text=white]{D} \\

    MM-MATH \citep{sun2024mm} \FiveStarConvex & EMNLP Findings'24 & English & 5,929 & \encircle[fill=DarkBlue, text=white]{S} & \encircle[fill=myBlue, text=white]{M} \hspace{2pt}\encircle[fill=myYellowv2, text=white]{H} & Discriminative & Closed/Open & \encircle[fill=DarkRed, text=white]{S} \hspace{2pt}\encircle[fill=DarkRed, text=white]{D} \\

    Scieval \citep{sun2024scieval} & AAAI'24 & English & 15,901 & \encircle[fill=DarkBlue, text=white]{S} \hspace{2pt}\encircle[fill=DarkBlue, text=white]{P} & \encircle[fill=MyRedv2, text=white]{H} & Both & Closed/Open &  \encircle[fill=DarkRed, text=white]{S} \\

    ArqMATH \citep{satpute2024can} & SIGIR'24 & English & 450 & \encircle[fill=DarkBlue, text=white]{P} & \encircle[fill=myOrangev2, text=white]{U} & Generative & Closed/Open/Math & \encircle[fill=DarkRed, text=white]{S}  \\

    IsoBench \citep{fu2024isobench} \FiveStarConvex & COLM'24 & English & 1,887 & \encircle[fill=DarkBlue, text=white]{S} & \encircle[fill=MyGreenv2, text=white]{E} \hspace{2pt}\encircle[fill=myBlue, text=white]{M} \hspace{2pt}\encircle[fill=myYellowv2, text=white]{H} \hspace{2pt}\encircle[fill=myOrangev2, text=white]{U} & Discriminative & Closed/Open & \encircle[fill=DarkRed, text=white]{S} \\

    $\text{MMMU-Pro}_{\text{math}}$ \citep{yue2024mmmu} \FiveStarConvex & arXiv'24 & English & 60 & \encircle[fill=DarkBlue, text=white]{S} & \encircle[fill=myOrangev2, text=white]{U} & Discriminative & Closed/Open & \encircle[fill=DarkRed, text=white]{S} \\    

    MathOdyssey \citep{fang2024mathodyssey} & arXiv'24 & English & 387 & \encircle[fill=DarkBlue, text=white]{S} & \encircle[fill=myYellowv2, text=white]{H} \hspace{2pt}\encircle[fill=myOrangev2, text=white]{U} \hspace{2pt}\encircle[fill=myViolet, text=white]{C} & Both & Closed/Open/Math &  \encircle[fill=DarkRed, text=white]{S}  \\

    MathScape \citep{zhou2024mathscape} \FiveStarConvex & arXiv'24 & Chinese & 1,325 & \encircle[fill=DarkBlue, text=white]{S} & \encircle[fill=MyGreenv2, text=white]{E} \hspace{2pt}\encircle[fill=myBlue, text=white]{M} \hspace{2pt}\encircle[fill=myYellowv2, text=white]{H} & Generative & Closed/Open &  \encircle[fill=DarkRed, text=white]{S}  \\

    U-Math \citep{ko2024umath} \FiveStarConvex & arXiv'24 & English & 1,100 & \encircle[fill=DarkBlue, text=white]{S} & \encircle[fill=myOrangev2, text=white]{U} & Discriminative & Closed/Open/Math & \encircle[fill=DarkRed, text=white]{S} \hspace{2pt}\encircle[fill=DarkRed, text=white]{D} \\

    MathHay \citep{wang2024mathhay} & arXiv'24 & English & 673 & \encircle[fill=DarkBlue, text=white]{S} \hspace{2pt}\encircle[fill=DarkBlue, text=white]{P} & \encircle[fill=MyRedv2, text=white]{H} & Both & Closed/Open &  \encircle[fill=DarkRed, text=white]{S}  \\

    FaultyMath \citep{rahman2024blind} \FiveStarConvex & arXiv'24 & English & 363 & \encircle[fill=DarkBlue, text=white]{G} & \encircle[fill=MyGreenv2, text=white]{E} \hspace{2pt}\encircle[fill=myBlue, text=white]{M} \hspace{2pt}\encircle[fill=myYellowv2, text=white]{H} & Discriminative & Closed/Open/Math & \encircle[fill=DarkRed, text=white]{D}  \\

    MathChat \citep{liang2024mathchat} & arXiv'24 & English & 1,319 & \encircle[fill=DarkBlue, text=white]{P} & \encircle[fill=MyGreenv2, text=white]{E} & Both & Closed/Open/Math & \encircle[fill=DarkRed, text=white]{S} \hspace{2pt}\encircle[fill=DarkRed, text=white]{D} \hspace{2pt}\encircle[fill=DarkRed, text=white]{O} \\

    E-GSM \citep{xu2024can} & arXiv'24 & Chinese & 4,500 & \encircle[fill=DarkBlue, text=white]{P} & \encircle[fill=MyGreenv2, text=white]{E} & Both & Closed/Open/Math & \encircle[fill=DarkRed, text=white]{S} \hspace{2pt}\encircle[fill=DarkRed, text=white]{O} \\

    Tangram \citep{tang2024tangram} \FiveStarConvex & arXiv'24 & English & 4,320 & \encircle[fill=DarkBlue, text=white]{S} & \encircle[fill=MyGreenv2, text=white]{E} \hspace{2pt}\encircle[fill=myBlue, text=white]{M} \hspace{2pt}\encircle[fill=myYellowv2, text=white]{H} \hspace{2pt}\encircle[fill=myViolet, text=white]{C} & Discriminative & Closed/Open & \encircle[fill=DarkRed, text=white]{O} \\

    CMM-Math \citep{liu2024cmm} \FiveStarConvex & arXiv'24 & Chinese & 28,069 & \encircle[fill=DarkBlue, text=white]{S} & \encircle[fill=MyGreenv2, text=white]{E} \hspace{2pt}\encircle[fill=myBlue, text=white]{M} \hspace{2pt}\encircle[fill=myYellowv2, text=white]{H} & Both & Closed/Open/Math &  \encircle[fill=DarkRed, text=white]{S} \\

    CMMaTH \citep{li2024cmmath} \FiveStarConvex & arXiv'24 & English/Chinese & 23,856 & \encircle[fill=DarkBlue, text=white]{S} & \encircle[fill=MyGreenv2, text=white]{E} \hspace{2pt}\encircle[fill=myBlue, text=white]{M} \hspace{2pt}\encircle[fill=myYellowv2, text=white]{H} & Both & Closed/Open/Math &  \encircle[fill=DarkRed, text=white]{S}  \\

    EAGLE \citep{li2024eagle} \FiveStarConvex & arXiv'24 & English & 170,000 & \encircle[fill=DarkBlue, text=white]{P} & \encircle[fill=MyGreenv2, text=white]{E} \hspace{2pt}\encircle[fill=myBlue, text=white]{M} \hspace{2pt}\encircle[fill=myYellowv2, text=white]{H} & Discriminative & Closed/Open/Math & \encircle[fill=DarkRed, text=white]{S}  \\

    VisAidMath \citep{ma2024visaidmath} \FiveStarConvex & arXiv'24 & English & 1,200 & \encircle[fill=DarkBlue, text=white]{S} & \encircle[fill=myBlue, text=white]{M} \hspace{2pt}\encircle[fill=myYellowv2, text=white]{H} \hspace{2pt}\encircle[fill=myViolet, text=white]{C} & Discriminative & Closed/Open &  \encircle[fill=DarkRed, text=white]{S}  \\

    AutoGeo \citep{huang2024autogeo} \FiveStarConvex & arXiv'24 & English & 100,000 & \encircle[fill=DarkBlue, text=white]{S} & \encircle[fill=MyGreenv2, text=white]{E} \hspace{2pt}\encircle[fill=myBlue, text=white]{M} \hspace{2pt}\encircle[fill=myYellowv2, text=white]{H} \hspace{2pt}\encircle[fill=myOrangev2, text=white]{U} & Both & Closed/Open & \encircle[fill=DarkRed, text=white]{O} \\

    NTKEval \citep{guo2024learning} & arXiv'24 & English & 1,860 &  \encircle[fill=DarkBlue, text=white]{P} \hspace{2pt}\encircle[fill=DarkBlue, text=white]{G} & \encircle[fill=MyRedv2, text=white]{H} & Discriminative & Open &  \encircle[fill=DarkRed, text=white]{S}  \\

    Mamo \citep{huang2024mamo} & arXiv'24 & English & 1,209 & \encircle[fill=DarkBlue, text=white]{S} \hspace{2pt}\encircle[fill=DarkBlue, text=white]{G} &  \encircle[fill=myOrangev2, text=white]{U} & Generative & Closed/Open/Math & \encircle[fill=DarkRed, text=white]{O} \\

    RoMath \citep{cosma2024romath} & arXiv'24 & Romanian & 70,000 & \encircle[fill=DarkBlue, text=white]{S} & \encircle[fill=myBlue, text=white]{M} \hspace{2pt}\encircle[fill=myYellowv2, text=white]{H} \hspace{2pt}\encircle[fill=myViolet, text=white]{C} & Discriminative & Closed/Open/Math & \encircle[fill=DarkRed, text=white]{S} \\

    MaTT \citep{davoodi2024llms} & arXiv'24 & English & 1,958 & \encircle[fill=DarkBlue, text=white]{S} & \encircle[fill=myOrangev2, text=white]{U} & Discriminative & Closed/Open & \encircle[fill=DarkRed, text=white]{S} \\

    \citet{li2024common} & arXiv'24 & English & 15,000 & \encircle[fill=DarkBlue, text=white]{P} & \encircle[fill=MyGreenv2, text=white]{E} \hspace{2pt}\encircle[fill=myBlue, text=white]{M} \hspace{2pt}\encircle[fill=myYellowv2, text=white]{H} & Generative & Closed/Open/Math & \encircle[fill=DarkRed, text=white]{S}  \\

    PolyMATH \citep{gupta2024polymath} \FiveStarConvex & arXiv'24 & English & 5,000 & \encircle[fill=DarkBlue, text=white]{S} & \encircle[fill=myBlue, text=white]{M} \hspace{2pt}\encircle[fill=myYellowv2, text=white]{H} \hspace{2pt}\encircle[fill=myOrangev2, text=white]{U} & Discriminative & Closed/Open & \encircle[fill=DarkRed, text=white]{S}  \\

    SuperCLUE-Math6 \citep{xu2024superclue} & arXiv'24 & English/Chinese & 2,144 & \encircle[fill=DarkBlue, text=white]{S} & \encircle[fill=MyGreenv2, text=white]{E} & Generative & Closed/Open & \encircle[fill=DarkRed, text=white]{S}  \\

    TheoremQA \citep{chen2023theoremqa} & EMNLP'23 & English & 800 & \encircle[fill=DarkBlue, text=white]{S} & \encircle[fill=myOrangev2, text=white]{U} & Discriminative & Closed/Open & \encircle[fill=DarkRed, text=white]{S}  \\

    LILA \citep{mishra2022lila} & EMNLP'22 & English & 133,815 & \encircle[fill=DarkBlue, text=white]{P} & \encircle[fill=MyRedv2, text=white]{H}& Discriminative & Closed & \encircle[fill=DarkRed, text=white]{S}  \\

    GeoQA \citep{chen2021geoqa}  \FiveStarConvex & ACL'21 & Chinese & 4,998 & \encircle[fill=DarkBlue, text=white]{S} & \encircle[fill=myBlue, text=white]{M} & Discriminative & Open & \encircle[fill=DarkRed, text=white]{S}  \\

    MATH \citep{hendrycks2021measuring} & NeurIPS'21 & English & 12,500 & \encircle[fill=DarkBlue, text=white]{S} & \encircle[fill=myViolet, text=white]{C} & Discriminative & Closed & \encircle[fill=DarkRed, text=white]{S}  \\

    \bottomrule
    \end{tabular}
}
\caption{\textbf{Overview of LLM-based benchmarks for mathematical reasoning}. \FiveStarConvex refers to those designed to evaluate the multimodal mathematical setting. Different colors indicate different types for the following columns:
\\ \textbf{Source}: \encircle[fill=DarkBlue, text=white]{S} = \underline{S}elf-Sourced, \encircle[fill=DarkBlue, text=white]{P} = Collected from \underline{P}ublic Dataset, \encircle[fill=DarkBlue, text=white]{G} = \underline{G}enerated by LLM
\\ \textbf{Level}: \encircle[fill=MyGreenv2, text=white]{E} = \underline{E}lementary, \encircle[fill=myBlue, text=white]{M} = \underline{M}iddle School, \encircle[fill=myYellowv2, text=white]{H} = \underline{H}igh School, \encircle[fill=myOrangev2, text=white]{U} = \underline{U}niversity, \encircle[fill=myViolet, text=white]{C} = \underline{C}ompetition, \encircle[fill=MyRedv2, text=white]{H} = \underline{H}ybrid
\\ \textbf{Task}: \encircle[fill=DarkRed, text=white]{S} = Problem-\underline{S}olving, \encircle[fill=DarkRed, text=white]{D} = Error \underline{D}etection, \encircle[fill=DarkRed, text=white]{P} = \underline{P}roving, \encircle[fill=DarkRed, text=white]{O} = \underline{O}thers
}
\label{tab:eval_datasets}
\end{table*}

\begin{table*}[!t]
\tiny
\centering
\resizebox{\linewidth}{!}{
    \begin{tabular}{lcccccp{5.5cm}c} 
    \toprule
    \textbf{Math (M)LLMs} & \textbf{Organization} & \textbf{Release Date} & \textbf{Publication} & \textbf{Language}  & \textbf{Parameter Size} & \textbf{Evaluation Benchmarks} & \textbf{Open Source}    \\
    \midrule
    GPT-f  ~\citep{polu2020generative}& OpenAI & Sep 2020 & - & English & 160M/400M/700M & - & \ding{52} \\ 
    Hypertree Proof Search~\citep{lample2022hypertree} & Meta & Nov 2022 & NeurIPS'22 & English & - & miniF2F/Metamath & - \\ 
    Minerva~\citep{lewkowycz2022solving} & Google & Jun 2022 & NeurIPS'22 & English & 8B/62B/540B & MATH/MMLU-STEM/GSM8k & - \\ 
    JiuZhang 1.0 ~\citep{zhao2022jiuzhang} & RUC \& iFLYTEK & Jun 2022 & KDD'22 & English & 145M & - & \ding{52} \\ 
    GAIRMath-Abel ~\citep{gairmathabel}& Shanghai Jiaotong University & 2023 & - & English & 7B/13B/70B & GSM8K/MATH/MMLU/SVAMP/SCQ5K-English/MathQA & \ding{52} \\ 
    JiuZhang 2.0  ~\citep{zhao2023jiuzhang}& RUC \& iFLYTEK & 2023 & KDD ADS'23 & English & - & JCAG/JBAG (MathBERT/DART/JiuZhang) & \ding{52} \\ 
    KwaiYiiMath ~\citep{fu2023kwaiyiimath}& Kuaishou & Jan 2023 & - & English/Chinese & 13B & GSM8K/CMath/KMath & - \\ 
    MathCoder ~\citep{wang2023mathcoder}& CUHK & Jan 2023 & ICLR'24 & English & 7B/13B & GSM8K/MATH & \ding{52} \\ 
    Llemma ~\citep{azerbayev2023llemma}& Princeton University \& Eleuther AI & Jan 2023 & - & English & 7B/34B & MATH/GSM8k/MMLU-STEM/SAT/OCWCourse & \ding{52} \\ 
    Skywork-13B-Math~\citep{zeng2024skywork} \FiveStarConvex  & SkyworkAI & Jan 2023 & - & English & 7B/13B & GSM8K/CMATH/MATH & \ding{52} \\ 
    MathGPT  ~\citep{mathgpt}\FiveStarConvex & TAL Education Group & Aug 2023 & - & English/Chinese & 130B & CEval-Math/AGIEval-Math/APE5K/CMMLU-Math/GAOKAO-Math/Math401 & - \\ 
    WizardMath~\citep{luo2023wizardmath} & Microsoft & Aug 2023 & ICLR'25 & English & 7B/70B & GSM8K/MATH & \ding{52} \\ 
    MAmmoTH1~\citep{yue2023mammoth} & UWaterloo & Sep 2023 & ICLR'24 & English & 7B/13B/70B & GSM/MATH/MMLU-STEM/AQuA/NumGLUE & \ding{52} \\ 
    MathGLM ~\citep{yang2023gpt}& Tsinghua \& Zhipu.AI & Sep 2023 & - & English & 10M/100M/500M/2B(Arith.)\&335M/6B/10B (MWP) & BIG-bench/ Ape210K & \ding{52} \\ 
    MetaMath ~\citep{yu2023metamath}& Cambridge \& Huawei & Sep 2023 & - & English & 7B/13B/70B & GSM8k/MATH & \ding{52} \\ 
    DeepSeekMath ~\citep{shao2024deepseekmath}& DeepSeek AI & Jan 2024 & - & English & 7B & GSM8K/MATH/OCW/SAT/MMLU-STEM/CMATH/Gaokao-MathCloze/Gaokao-MathQA & \ding{52} \\ 
    InternLM2.5-StepProver ~\citep{wu2024internlm2} & Shanghai AI Lab & Jan 2024 & - & English/Chinese & 7B & miniF2F/Lean-Workbook-Plus/ProofNet/Putnam & \ding{52} \\ 
    ChatGLM-Math ~\citep{xu2024chatglm}& Zhipu.AI & Apr 2024 & - & English/Chinese  & 32B & MathUserEval/Ape210k/CMath/GSM8k/MATH/Hungarian & - \\ 
    Rho-Math ~\citep{lin2024rho}& Microsoft & Apr 2024 & - & English & 1B/7B & GSM8K/MATH/MMLU-STEM/SAT/SVAMP/ASDiv/MAWPS/TAB/MQA & \ding{52} \\ 
    DeepSeekProver-V1 ~\citep{xin2024deepseekv1}& DeepSeek AI & May 2024 & - & English & 7B & miniF2F/FIMO & - \\ 
    InternLM2-Math ~\citep{wu2024internlm2}& Shanghai AI Lab & May 2024 & - & English/Chinese & 1.8B/7B/20B/8x22B & MiniF2F-test/MATH/MATH-Python/GSM8K/MathBench-A/Hungary/ & \ding{52} \\ 
    JiuZhang 3.0 ~\citep{zhou2024jiuzhang3}& RUC \& iFLYTEK & May 2024 & NeurIPS'24 & English & 7B/8B & GSM8k/MATH/G-Hard/SVAMP/MAWPS/ASDiv/TabMWP & \ding{52} \\ 
    MAmmoTH2 ~\citep{yue2024mammoth2}& UWaterloo & May 2024 & - & English & 7B/8B & TheoremQA/MATH/GSM8K/GPQA/MMLU-STEM/BBH & \ding{52} \\ 
    Math-LLaVA~\citep{shi2024math} & NUS & Jun 2024 & EMNLP Finding'24 & English & 13B & MMMU/MATH-V/MathVista & \ding{52} \\ 
    
    Mathstral ~\citep{mathstral}& Mistral AI & Jul 2024 & - & English & 7B & MATH/GSM8K/GREMath/AMC2023/AIME2024/MathOdyssey & - \\ 
 
    DeepSeek-Prover-V1.5 ~\citep{xin2024deepseekp}& DeepSeek AI & Aug 2024 & - & English & 7B & miniF2F-test/ProofNet & \ding{52} \\ 
    Qwen2-Math ~\citep{qwen2math}& Alibaba & Aug 2024 & - & English/Chinese  & 1.5B/7B/72B & GSM8K/Math/MMLU-STEM/CMATH/GaoKaoMath Cloze/GaoKao Math QA & \ding{52} \\ 
    Qwen2-Math-Instruct ~\citep{qwen2math}& Alibaba & Aug 2024 & - & English/Chinese  & 1.5B/7B/72B & GSM8K/MATH/Minerva Math/GaoKao2023 En/Olympiad Bench/College Math/MMLU STEM/Gaokao/CMATH/CNMiddle School 24/AIME24/AMC23 & \ding{52} \\ 
     MathGLM-Vision ~\citep{yang2024mathglm} \FiveStarConvex & Tsinghua \& Zhipu.AI & Sep 2024 & - & English & 9B/19B/32B & MathVista/MathVista(GPS)/MathVerse/Math-Vision/MMMU/MathVL & - \\ 
    Math-LLM~\citep{liu2024cmm} \FiveStarConvex  & East China Normal University & Sep 2024 & - & Chinese  & 8.26B/7B/72B & CMM-Math/MathVista/Math-V & - \\ 
    Qwen2.5-Math~\citep{yang2024qwen25} & Alibaba & Sep 2024  & - & English/Chinese  & 1.5B/7B/72B & GSM8K/MATH/MMLU-STEM/CMATH/GaoKao Math & \ding{52}  \\ 
   
    Xwin-LM~\citep{ni2024xwin} & Microsoft & May 2024 & - & English & 7B/13B/70B & GSM8K/MATH & \ding{52} \\ 
    MathCoder2 ~\citep{lu2024mathcoder2bettermathreasoning}& CUHK & Nov 2024 & ICLR'25 & English & 7B & GSM8K/MATH/SAT-Math/OCW/MMLU-Math & \ding{52} \\ 
    math-specialized Gemini 1.5 Pro \FiveStarConvex  & Google & Not launched yet & - & English & - & MATH/AIME2024/Math Odyssey/HiddenMath/IMO Bench & - \\ 
    k0-math ~\citep{k0math}& Moonshot AI & Nov 2024 & - & English/Chinese & - & KAOYAN/MATH/AIME/OMNI-MATH/GAOKAO/ZHONGKAO & - \\ 
    Duolingo Math  ~\citep{duolingo} & Duolingo & 2024 & - & English & - & - & - \\ 
    Khanmigo  ~\citep{khanmigo}& Khan Academy & 2024 & - & English & - & - & - \\ 
    Squirrel LAM ~\citep{squirrelai} \FiveStarConvex & Squirrel Ai Learning & 2024 & - & Chinese & - & - & - \\
    \bottomrule
    \end{tabular}
}
\caption{\textbf{Overview of math-specific LLMs (sort by release date).} \FiveStarConvex refers to those designed to support the multimodal mathematical setting.
}
\label{tab:math-llm}
\end{table*}

\begin{figure*}[t!]
    \centering
    \includegraphics[width=\linewidth,scale=1.00]{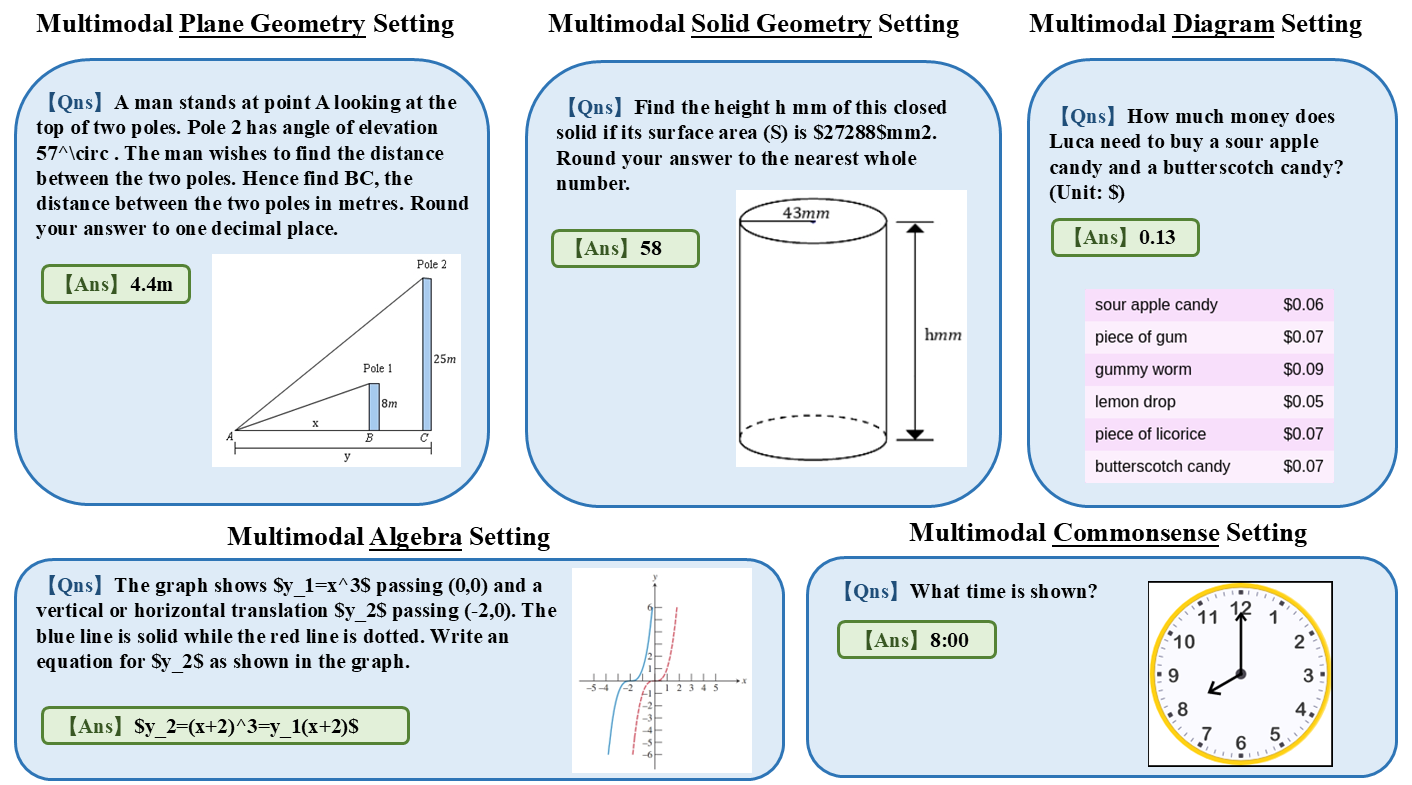}
    \caption{The illustration of diverse multimodal mathematical settings.}
    \label{fig:multimodal_cases}
\end{figure*}

\section{Summary of Benchmarks}
\label{app:benchmarks}
Table \ref{tab:eval_datasets} summarizes the LLM-based benchmarks for mathematical reasoning.

\section{Illustration of More Cases}
\label{app:cases}
Figure \ref{fig:multimodal_cases} illustrates the diverse multimodal cases of mathematical reasoning settings.
\subsection{Multimodal Plane Geometry Setting}
The Multimodal Plane Geometry Setting involves mathematical problems that require understanding and reasoning about 2D geometric relationships. These problems typically focus on fundamental geometric concepts, such as points, lines, angles, and triangles, often leveraging trigonometric principles like sine, cosine, or tangent. Visually, these questions are characterized by clear plane diagrams with labeled points, angles, and lengths. Students need to interpret these visuals to solve for unknown distances, angles, or other parameters. The defining feature here is the emphasis on 2D spatial relationships and the need to derive solutions from diagrammatic representations that combine measurements and geometry.

\subsection{Multimodal Solid Geometry Setting}
The Multimodal Solid Geometry Setting shifts the focus from 2D to 3D shapes and figures, such as cylinders, spheres, cubes, or cones. These questions often require students to compute surface area, volume, or height based on given measurements or constraints. Visually, these questions feature 3D diagrams with dimensions like radius, height, or length, typically annotated on the figure to help guide problem-solving. The main distinction is the incorporation of three-dimensional spatial reasoning and the need to analyze geometric properties of solids rather than flat, planar relationships. These tasks challenge students to bridge visual understanding with formulas involving multiple dimensions.

\subsection{Multimodal Diagram Setting}
In the Multimodal Diagram Setting, the problems revolve around interpreting visual data presented in the form of tables, charts, or diagrams. These tasks require students to extract numerical or categorical information and perform basic operations, such as addition, comparison, or selection. The visual components often include neatly organized tables, bar charts, or pie graphs, where the information is clearly labeled for accessibility. Unlike geometry-based problems, which require spatial reasoning, diagram settings focus on numerical literacy and the ability to synthesize information from structured visual data. This type highlights the integration of simple arithmetic and the comprehension of organized visual representations.

\subsection{Multimodal Algebra Setting}
The Multimodal Algebra Setting introduces problems that combine graphical representations and algebraic reasoning. These tasks often involve interpreting visual graphs, identifying equations, or understanding transformations such as translations or reflections. The visuals typically feature coordinate graphs with curves or lines, where solid and dotted lines may represent different functions or changes. Students are required to connect the visual graph to algebraic expressions, such as equations or transformations of functions. This type of question emphasizes the interplay between visual understanding (graph) and symbolic representation (algebra), making it distinct from purely numerical or geometric settings.

\subsection{Multimodal Commonsense Setting}
The Multimodal Commonsense Setting is characterized by problems that involve interpreting everyday visuals and applying logical reasoning. These questions present familiar objects, such as clocks, calendars, or real-world scenarios, where students must analyze the visual information to derive straightforward answers. Visually, these tasks feature clear and relatable imagery, like an analog clock with its hands pointing to a specific time. Unlike other types, commonsense settings rely less on abstract mathematical reasoning and more on practical interpretation of everyday visual cues. This setting highlights how mathematical understanding can intersect with routine, real-world observations.

\subsection{Summary}
In summary, the key differences among these types stem from their visual focus and cognitive demands. While plane and solid geometry emphasize spatial reasoning in 2D and 3D, respectively, diagram settings target numerical literacy through organized data. Algebra settings merge visual graphs with algebraic transformations, and commonsense settings leverage real-world visuals requiring practical logic. Each type uniquely integrates multimodal elements to challenge students across different mathematical skills.

\section{Details of Metrics}
\label{app:metrics}

\subsection{Discriminative Metrics}
Discriminative tasks refer to evaluation processes where the outputs are typically binary, such as "Yes" or "No". These tasks often include multiple-choice questions, fill-in-the-blank problems, or judgment assessments. The evaluation metrics focus on LLM's accuracy in specific task types and its ability to control biases.

\textbf{Accuracy (ACC):} It measures the proportion of correctly predicted outcomes. The value should be as high as possible.
\[
ACC = \frac{\sum_{1,m}x_i}{\sum_{1,n}y_j}
\]
Where $\boldsymbol{x}_{i}$ represents the correct output for the $\boldsymbol{i}$-th instance, $\boldsymbol{y}_{j}$ represents the $\boldsymbol{j}$-th instance, $m$ is the number of the correct instances and $n$ is the number of the total instances.

\textbf{Exact match:} It evaluates the congruence between the answers generated by LLM and the correct ones. Specifically, in cases where the answer produced LLM coincides with the reference answer, a score of 1 point will be assigned. Conversely, if there is any discrepancy between them except for bias, a score of 0 point will be given.

\textbf{\bm{$F_1$} score:} It combines two crucial aspects, namely precision and recall, in order to comprehensively assess the accuracy of LLM. It is calculated as :
\[F_1=2\times\frac{\mathrm{Precision}\times\mathrm{Recall}}{\mathrm{Precision}+\mathrm{Recall}}\]
The value of the $F_1$ score ranges from 0 to 1. A higher value of the $F_1$ score indicates better overall performance of LLM in terms of both precision and recall.

\textbf{Macro-\bm{$F_1$} score:} It calculates the $F_1$ score for each category separately and then takes the average of the $F_1$ scores of all categories, so as to obtain the overall performance of LLM on all categories.

\textbf{Round-r accuracy:} It is the proportion of correct answers given by a model on the question set Qr in round r. It is calculated as follows:
\[ACC_r(M)=\frac{\sum_{q\in Q_r}I[M(q)=g_t(q)]}{|Q_r|}\]
Here, $ACC_r(M)$ represents the accuracy of LLM $M$ on question set $Q_r$ in round $r$. $I$ is an indicator function. When the answer $M(q)$ given by $M$ for question $q$ is consistent with the true answer $g_t(q)$ of the question, the value of $I$ is 1; otherwise, it is 0. The symbol $\sum_{q\in Q_r}$ means summing over all questions in question set $Q_r$. $\left|Q_r\right|$ indicates the number of questions in question set $Q_r$.

\bm{$ACC_{step}:$} It is used to evaluate LLM's ability to identify the first step where an error occurs. The accuracy for identifying the first erroneous step is calculated as follows:
\[ACC_{step}=\frac{1}{N}\sum_{i=1}^N\mathbb{I}(S_{step,i}=G_{step,i})\]
Here, $N$ is the total number of samples. For the $i$-th sample, $S_{step,i}$ is the predicted step where the error occurs, and $G_{step,i}$ is the ground truth label for the first erroneous step. The indicator function $\mathbb{I}(\cdot)$ returns 1 if the predicted step matches the ground truth and 0 otherwise.

\bm{$ACC_{cate}:$} It is for assessing LLM's performance in categorizing the type of error. The accuracy for error categorization is defined by
\[ACC_{cate}=\frac{1}{N}\sum_{i=1}^N\mathbb{I}(C_{error,i}=G_{error,i})\]
Here, $N$ is the total number of samples. For the $i$-th sample, $C_{error,i}$ is the predicted error category, and $G_{error,i}$ is the ground truth label for the error category. The indicator function $\mathbb{I}(\cdot)$ has the same meaning as in the previous metric, returning 1 if the predicted error category matches the ground truth and 0 otherwise.

\textbf{The skill success rate:} It measures the proportion of a model correctly applying major skills in problem-solving. It's calculated by analyzing test questions and determining correct use of major skills, then finding the ratio to total questions. For example, in triangle area calculation, checking use of the area formula. Similarly, \textbf{the secondary skill success rate} focuses on the proportion of correct application of secondary skills like understanding graphic properties and unit conversion, calculated by analyzing problem-solving and finding the ratio to total questions.

\textbf{The False Positive Rate (FPR):} It is the proportion of cases where the evaluation LLM misjudges an incorrect answer as a correct one. A low FPR indicates that LLM rarely misjudges incorrect student answers as correct.

\textbf{The False Negative Rate (FNR):} It is the proportion of cases where the evaluation LLM misjudges a correct answer as an incorrect one. A low FNR indicates that LLM is relatively accurate in correctly determining whether a student's answer is correct.

\textbf{Mean Squared Error (MSE):} It is a metric that measures the average of the squares of the differences between the LLM's predicted values and the actual true values. It is calculated as: 
\[\begin{aligned}MSE=\frac{1}{n}\sum_{i=1}^n(y_i-\hat{y}_i)^2\end{aligned}\]
Here, $n$ represents the number of samples. For the $i$-th sample, $y_i$ is the true value and $\hat{y}_i$ is the predicted value by LLM. The summation symbol $\sum_{i=1}^n$ means summing up the squared differences for all $n$ samples. Dividing by $n$ gives the average squared difference, which is the MSE. MSE should be as low as possible.

\textbf{Average-Case Accuracy (\bm{$A_{avg}$}):} This metric evaluates the average accuracy of LLM across all variants of a seed question. It is calculated as the proportion of correct answers across all variants and seed questions. The formula is:
\[A_{avg}=\frac{1}{N}\sum_{i=1}^N\frac{1}{M}\sum_{j=1}^MI[\mathrm{Ans}(i,j)=\mathrm{GT}(i,j)]\]
where $N$ is the total number of seed questions, $M$ is the number of variants per seed question, and $I[\mathrm{Ans}(i,j)=\mathrm{GT}(i,j)]$ checks if the answer matches the ground truth.

\textbf{Worst-Case Accuracy (\bm{$A_{wst}$}): }This evaluates the worst-case performance by considering the minimum accuracy across all variants of a seed question. It reflects the robustness of LLM against challenging variations. The formula is:
\[A_{wst}=\frac{1}{N}\sum_{i=1}^N\min_{j\in[1,M]}I[\mathrm{Ans}(i,j)=\mathrm{GT}(i,j)]\]

\subsection{Generative Metrics}
Generative tasks involve evaluating the content generated by LLM, typically encompassing free-form answers and responses to open-ended questions. These tasks focus primarily on assessing the extent of hallucinations in the generated content, especially when the content is not faithful to the given images. Evaluating generative tasks often requires more complex metrics, such as CHAIR and Faithscore, which measure hallucinations across different categories, including objects, attributes, and relationships within the generated content. These metrics provide a nuanced understanding of the fidelity and reliability of MLLMs in producing content aligned with the visual and textual inputs.

\textbf{Reasoning Robustness (RR):} This metric measures the relative robustness of LLM by comparing the worst-case performance to the average-case performance. The formula is:
\[RR=\frac{A_{wst}}{A_{avg}}\]

\textbf{Repetition Consistency (RC):} This evaluates the consistency of LLM's responses across repeated queries for the same question variant. It helps distinguish between variability due to randomness and systematic errors. The formula is:
\[RC(i,j)=\frac{1}{K}\sum_{k=1}^KI[\mathrm{Ans}_k(i,j)=\mathrm{Ans}(i,j)]\]
where $K$ is the number of repetitions.

\textbf{OpenCompass Scoring:} It is a comprehensive evaluation framework that leverages the OpenCompass platform to assess the generative capabilities of LLM across multiple dimensions. Perplexity (PPL) evaluates the naturalness and fluency of generated text, with lower scores indicating greater model confidence and the ability to produce contextually coherent sequences. Simultaneously, CircularEval assesses the robustness and consistency of LLM in multiple-choice scenarios by evaluating its performance across $N$ random permutations of the options in an $N$-option question. A question is deemed correctly answered only if LLM provides the correct response for all permutations, highlighting its ability to handle randomized inputs reliably.

\textbf{Bilingual Evaluation Understudy (BLEU):} It evaluates the quality of text generation by measuring n-gram overlap between generated and reference texts, focusing on precision and brevity. Its formula is:
\[\mathrm{BLEU}=\mathrm{BP}\cdot\exp\left(\sum_{n=1}^Nw_n\log p_n\right)\]
where BP is the brevity penalty, calculated as 1 if $c>r$ , or $\exp(1-r/c)$ if $c\leq r$, with $c$ and $r$ representing the lengths of the generated and reference texts, respectively. $w_n$ denotes n-gram weights (typically uniform), and $p_n$ is the precision of n-grams of size $n$. BLEU scores range from 0 to 1 (often expressed as percentages, 0-100\%), with higher scores indicating greater similarity between the generated and reference texts. 

\textbf{Recall-Oriented Understudy for Gisting Evaluation-L (ROUGE-L):} It evaluates the quality of generated text by measuring its similarity to reference text, focusing on sequence alignment and structural consistency through the Longest Common Subsequence (LCS). It calculates recall as the proportion of the LCS length relative to the reference text length. The formula of recall is:
\[R=\frac{\mathrm{LCS}(\text{Generated, Reference})}{\mathrm{Length}(\mathrm{Reference})}\]
It also calculates precision as the proportion of the LCS length relative to the generated text length. The formula is:
\[P=\frac{\mathrm{LCS}(\text{Generated, Reference})}{\mathrm{Length}(\mathrm{Generated})}\]
The $F_1$ score is a harmonic mean of precision and recall, expressed as:
\[F_1=\frac{(1+\beta^2)\cdot P\cdot R}{\beta^2\cdot P+R}\]
where $\beta$ (commonly set to 1) controls the weighting of recall and precision. ROUGE-L scores range from 0 to 1, with higher scores indicating greater similarity between the generated and reference texts. 

\textbf{Consensus-based Image Description Evaluation (CIDEr):} It is designed for image description tasks, measuring the semantic relevance of generated descriptions by calculating the TF-IDF weighted n-gram similarity with reference descriptions. The formula is:
\[CIDEr_n(c_i,S_i)=\frac{1}{m}\sum_{j=1}^m\frac{g^n(c_i)\cdot g^n(s_{ij})}{||g^n(c_i)||\cdot||g^n(s_{ij})||}\]
\[CIDEr(c_i,S_i)=\sum_{n=1}^Nw_nCIDEr_n(c_i,S_i)\]
Here, $c_{i}$ is the candidate description, $S_i=\{s_{i1},s_{i2},\ldots,s_{im}\}$ is the set of reference descriptions, and $m$ is the number of references. $g^n(c_i)$ and $g^n(s_{ij})$ are the TF-IDF weighted n-gram vectors for the candidate and reference descriptions, with $||g^n(c_i)||$ and $||g^n(s_{ij})||$ being their magnitudes. $w_n$ is the weight for n-grams of different lengths, usually $w_n=1/N$, where $N$ is the maximum n-gram length. Scores range from 0 to 10, with higher scores indicating stronger alignment between candidate and reference descriptions.

\textbf{Mathematical Symbol Similarity:} This metric measures the similarity between the correct steps in a reasoning process and the steps generated by LLM, using symbolic computation software to perform the evaluation.

\textbf{GPT Scoring:} This metric evaluates the generated content based on scores assigned by GPT or other language models, focusing on the linguistic coherence and logical consistency of the text.

\textbf{Context Length Generalization Efficacy (CoLeG-E):} It is a metric used to measure LLM's consistency in answering variations of the same question across different context lengths. It is defined as:
\[CoLeG\mathrm{-}E(M)=\frac{\sum_{q\in Q_R}\left[\bigwedge_{r=1}^RI[M(q^r)=gt(q^r)]\right]}{|Q_R|}\]
where $Q_R$ represents the set of all questions under evaluation, and $q^{r}$ refers to the $r$-th variation of a question $q$, corresponding to a specific context length. $M(q^{r})$ is LLM's predicted answer for the $r$-th variation, while $M(q^{r})$ denotes the ground truth answer. The indicator function $I[\cdot]$ equals 1 if LLM's answer matches the ground truth, and 0 otherwise. The logical AND operator $\Lambda_{r=1}^R$ ensures that the model must answer all variations of a question correctly for that question to be considered correctly answered. 

\textbf{Context Length Generalization Robustness (CoLeG-R):} It measures LLM's robustness to context length expansion by quantifying the relative drop in accuracy from initial to extended questions. It is defined as:
\[CoLeG\mathrm{-}R(M)=1-\frac{\mathrm{ACC}_0(M)-\mathrm{ACC}_R(M)}{\mathrm{ACC}_0(M)}\]
Here, $\mathrm{ACC}_0(M)$ is the LLM's accuracy on the initial set of shorter-context questions $Q_0$, and $\mathrm{ACC}_R(M)$ is its accuracy on the extended longer-context questions $Q_R$. Higher CoLeG-R values indicate better robustness, with less performance degradation across context lengths.

\textbf{Performance Drop Rate (PDR):} This metric measures the relative decline in model performance when transitioning from the original dataset to the perturbed dataset. It is defined as:
\[\mathrm{PDR}=1-\frac{\sum_{(x,y)\in D_a}I[\mathrm{LLM}(x),y]/|D_a|}{\sum_{(x,y)\in D}I[\mathrm{LLM}(x),y]/|D|}\]
where D is the original dataset and $D_a$ is the perturbed dataset. $I[\mathrm{LLM}(x),y]$ is an indicator function that checks if the LLM’s output matches the ground truth $y$.

\textbf{Accurately Solved Pairs (ASP):} ASP measures the percentage of seed questions and their perturbed variations that are both correctly answered by LLM. It is defined as:
\[\mathrm{ASP}=\frac{\sum_{x,y;x^{\prime},y^{\prime}}I[\mathrm{LLM}(x),y]\cdot I[\mathrm{LLM}(x^{\prime},y^{\prime})]}{N\cdot|D|}\]
where $x$ and $x^{\prime}$ are a seed question and its variation, respectively. $N$ is the number of perturbations per question. $\left|D\right|$ is the total number of seed questions.

\textbf{Mean Average Precision (mAP):} It is a metric that evaluates LLM's ability to rank relevant answers higher in its output list for a given query. It is defined as:
\[mAP=\frac{1}{|Q|}\sum_{q\in Q}AP(q)\]
\[AP(q)=\frac{1}{m}\sum_{k=1}^mP(k)\]
\[P(k)=\frac{\text{\# relevant ans retrieved up to position }k}{k}\]
Here, $Q$ represents the set of all queries in the dataset. $AP(q)$ is the Average Precision for query $q$, calculated as the mean of the precision values $P(k)$ at ranks where relevant answers appear.$P(k)$ is the precision at rank $k$, representing the proportion of relevant answers retrieved up to position $k$. $m$ is the total number of relevant answers for query $q$.

\textbf{Training Set Coverage (TSC):} It measures how effectively LLM has learned to generate correct solutions for tasks similar to those in its training set. TSC is particularly useful in cross-domain or cross-modal tasks, where it assesses LLM's ability to generalize learned patterns to problems aligned with its training data. Higher TSC scores indicate better learning and consistency, while lower scores suggest insufficient training or overfitting.

\textbf{Pass@N:} This metric measures the likelihood of LLM generating at least one correct solution within $N$ attempts for a given problem. Formally:
\[Pass@N=\mathbb{E}_{\mathrm{Problems}}[\min(c,1)]\]
where $c$ represents the number of correct answers out of $N$ responses. A higher Pass@N indicates a greater chance of producing a correct answer in multiple attempts, reflecting LLM's potential capability.

\textbf{PassRatio@N:} This metric calculates the proportion of correct answers among $N$ generated responses for a given problem. It is defined as:
\[PassRatio@N=\mathbb{E}_{\mathrm{Problems}}\left[\frac{c}{N}\right]\]
where $c$ is the count of correct answers. This metric reflects LLM's stability in consistently generating correct answers. It can be considered analogous to Pass@1 but offers reduced variance.

\section{Summary of Methods}
\label{app:methods}
Table \ref{tab:method} summarizes the LLM-based methods for mathematical reasoning.

\begin{table*}[!t]
\tiny
\centering
\resizebox{\linewidth}{!}{
    \begin{tabular}{lcp{4.5cm}ccccc} 
    \toprule
    \textbf{Methods} & \textbf{Venue} & \textbf{Evaluated Math Dataset(s)} & \textbf{Task(s)}  & \textbf{Scope(s)} & \textbf{LLM as Enhancer} & \textbf{LLM as Reasoner} & \textbf{LLM as Planner}   \\
    \midrule

    MathAgent \citep{yan2025mathagent} \FiveStarConvex & ACL'25& ErrorRadar & \encircle[fill=DarkRed, text=white]{D} & \encircle[fill=myOrangev2, text=white]{M} &  &  & \ding{52}\\
    
    MAVIS \citep{zhang2024mavis} \FiveStarConvex & ICLR'25& MathVerse/GeoQA/MathVista/MMMU/MathVision & \encircle[fill=DarkRed, text=white]{S} & \encircle[fill=myOrangev2, text=white]{M} & \ding{52} & \ding{52} & \\

    TVM \citep{lee2024token} & ICLR'25& GSM8K/MATH & \encircle[fill=DarkRed, text=white]{S} & \encircle[fill=myBlue, text=white]{A} &  & \ding{52} & \\
    
    MathCoder2 \citep {lu2024mathcoder2bettermathreasoning} & ICLR'25& GSM8K/MATH/SAT-Math/OCW/MMLU-Math & \encircle[fill=DarkRed, text=white]{S}\hspace{4pt}\encircle[fill=DarkRed, text=white]{P} & \encircle[fill=myOrangev2, text=white]{M} & \ding{52} & \ding{52}&\\
    
    \citet{xiong2024building}& ICLR'25& GSM8K/MATH& \encircle[fill=DarkRed, text=white]{S}\hspace{4pt}\encircle[fill=DarkRed, text=white]{P} & \encircle[fill=myBlue, text=white]{A}& & \ding{52}& \ding{52}\\

    TSMC \citep{feng2024step}& ICLR'25& GSM8K/MATH500& \encircle[fill=DarkRed, text=white]{S}\hspace{4pt}\encircle[fill=DarkRed, text=white]{P}  & \encircle[fill=myBlue, text=white]{A}& & \ding{52}&\\
    
    AlphaGeometry \citep{trinh2024solving} \FiveStarConvex & Nature'24  & IMO-AG-30 & \encircle[fill=DarkRed, text=white]{S}\hspace{4pt}\encircle[fill=DarkRed, text=white]{P}  & \encircle[fill=MyGreenv2, text=white]{G}& \ding{52} & \ding{52} &    \\ 
    
    Masked Thought \citep{chen2024masked}  & ACL'24  & GSM8K/MATH/GSM8K-RFT/MetaMathQA/MathInstruct & \encircle[fill=DarkRed, text=white]{S}  & \encircle[fill=myBlue, text=white]{A}& \ding{52} & \ding{52} &    \\
    
     MathGenie \citep{lu2024mathgenie}& ACL'24& GSM8K/MATH/SVAMP/Simuleq/Mathematics& \encircle[fill=DarkRed, text=white]{S}  & \encircle[fill=myBlue, text=white]{A}& \ding{52}& \ding{52}&\\
     
     MATH-SHEPHERD \citep{wang2024math}& ACL'24& GSM8K/MATH& \encircle[fill=DarkRed, text=white]{S}\hspace{4pt}\encircle[fill=DarkRed, text=white]{P} & \encircle[fill=myBlue, text=white]{A}& \ding{52}& \ding{52}&\\
     
     SEGO \citep{zhao2023sego}& ACL'24 & GSM8K/MATH & \encircle[fill=DarkRed, text=white]{S}\hspace{4pt}\encircle[fill=DarkRed, text=white]{P} & \encircle[fill=myBlue, text=white]{A}& \ding{52} & \ding{52} & \\
     
     \citet{deng2023towards}& ACL Workshop'24& GSM8K/SVAMP/MultiArith/MathQA/CSQA& \encircle[fill=DarkRed, text=white]{S}  & \encircle[fill=myBlue, text=white]{A}& & \ding{52}&\\
     
     MathCoder \citep
    {wang2023mathcoder} & ICLR'24& GSM8K/MATH& \encircle[fill=DarkRed, text=white]{S}\hspace{4pt}\encircle[fill=DarkRed, text=white]{P} & \encircle[fill=myBlue, text=white]{A}& & \ding{52}&\\
    
     ToRA \citep{gou2023tora}& ICLR'24& GSM8K/MATH& \encircle[fill=DarkRed, text=white]{S}\hspace{4pt}\encircle[fill=DarkRed, text=white]{P} & \encircle[fill=myBlue, text=white]{A}& & \ding{52}&\ding{52}\\
     
      Visual Sketchpad \citep{hu2024visual} \FiveStarConvex & NeurIPS'24& Geometry3K/ IsoBench& \encircle[fill=DarkRed, text=white]{S}& \encircle[fill=MyGreenv2, text=white]{G}& & &\ding{52}\\

      JiuZhang 3.0 \citep{zhou2024jiuzhang3}& NeurIPS'24& GSM8K/MATH/SVAMP/ASDiv/MAWPS/CARP& \encircle[fill=DarkRed, text=white]{S}\hspace{4pt}\encircle[fill=DarkRed, text=white]{P} & \encircle[fill=myBlue, text=white]{A}& \ding{52}& \ding{52}&\\
      
      Minimo \citep{poesia2024learning}& NeurIPS'24& -& \encircle[fill=DarkRed, text=white]{P}& \encircle[fill=myBlue, text=white]{A}& & \ding{52}&\\

      DART-Math \citep{tong2024dart}& NeurIPS'24& MATH/GSM8K/College/DM/Olympiad/Theorem& \encircle[fill=DarkRed, text=white]{S}\hspace{4pt}\encircle[fill=DarkRed, text=white]{P} & \encircle[fill=myBlue, text=white]{A}& \ding{52}& \ding{52}&\\
      
     \citet{li2024neuro} & NeurIPS'24& GSM8K/MATH & \encircle[fill=DarkRed, text=white]{S}  & \encircle[fill=myBlue, text=white]{A} & \ding{52} & &\\

     MACM \citep{lei2024macm} & NeurIPS'24 & MATH & \encircle[fill=DarkRed, text=white]{S}  & \encircle[fill=myBlue, text=white]{A} & & \ding{52} &\\

     \citet{sinha2024wu} \FiveStarConvex & NeurIPS Workshop'24& IMO-AG-30 & \encircle[fill=DarkRed, text=white]{S}\hspace{4pt}\encircle[fill=DarkRed, text=white]{P}  & \encircle[fill=MyGreenv2, text=white]{G}& & \ding{52}&\\
     
     SBIRAG \citep{dixit2024sbi}& NeurIPS Workshop'24& GSM8K& \encircle[fill=DarkRed, text=white]{S}  & \encircle[fill=myBlue, text=white]{A}& & \ding{52}&\\
     
     MathScale \citep{tang2024mathscale}& ICML'24& GSM8K/MATH/CollegeMath& \encircle[fill=DarkRed, text=white]{S}  & \encircle[fill=myBlue, text=white]{A}& \ding{52} & &\\
     
     VerityMath \citep{han2023veritymath}& ICML Workshop'24& GSM8K& \encircle[fill=DarkRed, text=white]{S}  & \encircle[fill=myBlue, text=white]{A}& & \ding{52}&\\
     
     RefAug \citep{zhang2024learn}& EMNLP'24& GSM8K/MATH/Mathematics/MAWPS/ SVAMP/MMLU-Math/SAT-Math/MathChat-FQA/MathChat-EC/MINI-Math& \encircle[fill=DarkRed, text=white]{S}\hspace{4pt}\encircle[fill=DarkRed, text=white]{D}\hspace{4pt}\encircle[fill=DarkRed, text=white]{P}& \encircle[fill=myBlue, text=white]{A}& \ding{52}& \ding{52}&\\
     
     Math-LLaVA  \citep{shi2024math} \FiveStarConvex & EMNLP Findings'24& MathVista/Math-V& \encircle[fill=DarkRed, text=white]{S}\hspace{4pt}\encircle[fill=DarkRed, text=white]{P} & \encircle[fill=myOrangev2, text=white]{M}& \ding{52} & \ding{52}&\\
     
 COPRA \citep{thakur2024context}& COLM'24& miniF2F-test& \encircle[fill=DarkRed, text=white]{S}& \encircle[fill=myBlue, text=white]{A}& & &\ding{52}\\
 PRP \citep{wu2024get}& AAAI'24& MAWPS/ASDivA/Math23k/SVAMP/Un-biasedMWP& \encircle[fill=DarkRed, text=white]{S}& \encircle[fill=myBlue, text=white]{A}& & \ding{52}&\\
 PERC \citep{jin2024using}& L@S'24& PERC& \encircle[fill=DarkRed, text=white]{S}& \encircle[fill=myBlue, text=white]{A}& & \ding{52}&\\
 Math-PUMA 
 \citep{zhuang2024math} \FiveStarConvex & arXiv'24& MathVerse/MathVista/WE-MATH& \encircle[fill=DarkRed, text=white]{S}\hspace{4pt}\encircle[fill=DarkRed, text=white]{P} & \encircle[fill=myOrangev2, text=white]{M}& & \ding{52}&\\
 MultiMath \citep{peng2024multimath} \FiveStarConvex & arXiv'24& MathVista/MathVerse/MultiMath-300K& \encircle[fill=DarkRed, text=white]{S}\hspace{4pt}\encircle[fill=DarkRed, text=white]{P} & \encircle[fill=myOrangev2, text=white]{M}& & \ding{52}&\\
 MathAttack \citep{zhou2024mathattack}& arXiv'24& GSM8K/MultiAirth& \encircle[fill=DarkRed, text=white]{S}  & \encircle[fill=myBlue, text=white]{A}& & \ding{52}&\\
 MinT \citep{liang2023mint}& arXiv'24& GSM8K/MathQA/CM17k/Ape210k& \encircle[fill=DarkRed, text=white]{S}  & \encircle[fill=myBlue, text=white]{A}& & \ding{52}&\\
 DotaMath \citep{li2024dotamath}& arXiv'24& GSM8K/MATH/Mathematics/SVAMP/TabMWP/ASDiv& \encircle[fill=DarkRed, text=white]{S}  & \encircle[fill=myBlue, text=white]{A}& \ding{52}& \ding{52}&\\
 DFE-GPS \citep{zhang2024diagram}& arXiv'24& FORMALGEO7k& \encircle[fill=DarkRed, text=white]{S}& \encircle[fill=MyGreenv2, text=white]{G}& \ding{52}& \ding{52}&\\
 
 PGPSNet-v2 \citep{zhang2024fuse} \FiveStarConvex & arXiv'24& Geometry3K/PGPS9K& \encircle[fill=DarkRed, text=white]{S}& \encircle[fill=MyGreenv2, text=white]{G}\hspace{4pt}\encircle[fill=myYellowv2, text=white]{D}& \ding{52}& \ding{52}&\\
 LLaMA-Berry \citep{zhang2024llama}& arXiv'24& GSM8K/MATH/GaoKao2023En/OlympiadBench/College Math/MMLU STEM& \encircle[fill=DarkRed, text=white]{S}\hspace{4pt}\encircle[fill=DarkRed, text=white]{P} & \encircle[fill=myBlue, text=white]{A}& & \ding{52}&\\
 Skywork-Math \citep{zeng2024skywork} \FiveStarConvex & arXiv'24& GSM8K/MATH& \encircle[fill=DarkRed, text=white]{S}\hspace{4pt}\encircle[fill=DarkRed, text=white]{P} & \encircle[fill=myBlue, text=white]{A}& \ding{52}& \ding{52}&\\
 SIaM \citep{yu2024siam}& arXiv'24& GSM8K/CMATH& \encircle[fill=DarkRed, text=white]{S}\hspace{4pt}\encircle[fill=DarkRed, text=white]{P} & \encircle[fill=myBlue, text=white]{A}& & \ding{52}&\\
 InternLM-Math \citep{ying2024internlm}& arXiv'24& GSM8K/MATH& \encircle[fill=DarkRed, text=white]{S}\hspace{4pt}\encircle[fill=DarkRed, text=white]{P} & \encircle[fill=myBlue, text=white]{A}& & \ding{52}&\\
 MathGLM-Vision \citep{yang2024mathglm} \FiveStarConvex & arXiv'24& MathVista/MathVerse/MathVision& \encircle[fill=DarkRed, text=white]{S}\hspace{4pt}\encircle[fill=DarkRed, text=white]{P} & \encircle[fill=myOrangev2, text=white]{M}& \ding{52}& \ding{52}&\\
 Qwen2.5-Math \citep{yang2024qwen25} \FiveStarConvex & arXiv'24& GSM8K/MATH/MMLU-STEM/CMATH/GaoKao-Math-Cloze/GaoKao-Math-QA& \encircle[fill=DarkRed, text=white]{S}\hspace{4pt}\encircle[fill=DarkRed, text=white]{P} & \encircle[fill=myBlue, text=white]{A}& \ding{52}& \ding{52}&\\
S3c-Math \citep{yan2024s}& arXiv'24& GSM8K/MATH/SVAMP/Mathematics& \encircle[fill=DarkRed, text=white]{S}\hspace{4pt}\encircle[fill=DarkRed, text=white]{P} & \encircle[fill=myBlue, text=white]{A}& \ding{52}& \ding{52}&\\
 SIRP \citep
{wu2024progress}& arXiv'24& CSQA/GSM8K/MATH/MBPP& \encircle[fill=DarkRed, text=white]{S}\hspace{4pt}\encircle[fill=DarkRed, text=white]{P} & \encircle[fill=myBlue, text=white]{A}& & \ding{52}&\\
 AIPS \citep{wei2024proving}& arXiv'24& MO-INT-20& \encircle[fill=DarkRed, text=white]{S}& \encircle[fill=MyGreenv2, text=white]{G}& & \ding{52}&\\

 DeepSeekMath \citep{shao2024deepseekmath}& arXiv'24& GSM8K/MATH/OCW/SAT/MMLU STEM/CMATH/Gaokao MathCloze/Gaokao MathQA& \encircle[fill=DarkRed, text=white]{S}  & \encircle[fill=myBlue, text=white]{A}& & \ding{52}&\\
 MMIQC \citep{liu2024augmenting}& arXiv'24& MATH/MMIQC& \encircle[fill=DarkRed, text=white]{S}  & \encircle[fill=myBlue, text=white]{A}& \ding{52}& \ding{52}&\\
 LANS \citep{li2023lans} \FiveStarConvex & arXiv'24& Geometry3K/PGPS9K& \encircle[fill=DarkRed, text=white]{S}& \encircle[fill=MyGreenv2, text=white]{G}\hspace{4pt}\encircle[fill=myYellowv2, text=white]{D}& & \ding{52}&\\
 VCAR \citep{jia2024describe} \FiveStarConvex & arXiv'24& MathVista/MathVerse& \encircle[fill=DarkRed, text=white]{S}  & \encircle[fill=myOrangev2, text=white]{M}& & \ding{52}&\\
 KPDDS \citep{huang2024key}& arXiv'24& GSM8k/MATH/SVAMP/TabMWP/ASDiv/MAWPS& \encircle[fill=DarkRed, text=white]{S}  & \encircle[fill=myBlue, text=white]{A}& \ding{52}& \ding{52}&\\
 HGR \citep{huang2024hologram} \FiveStarConvex & arXiv'24& Geometry3K
& \encircle[fill=DarkRed, text=white]{S}  & \encircle[fill=MyGreenv2, text=white]{G}& \ding{52}& \ding{52}&\\
 InfiMM-Math \citep{han2024infimm} \FiveStarConvex & arXiv'24& GSM8K/MMLU/MathVerse/We-Math& \encircle[fill=DarkRed, text=white]{S}  & \encircle[fill=myBlue, text=white]{A}& \ding{52}& &\\
 CoSC \citep{han2024infimm} & arXiv'24& GSM8K/MATH& \encircle[fill=DarkRed, text=white]{S}\hspace{4pt}\encircle[fill=DarkRed, text=white]{P}  & \encircle[fill=myBlue, text=white]{A}& \ding{52}& \ding{52}&\\

 SICCV \citep{liang2024improving}& arXiv'24& GSM8k/MATH500& \encircle[fill=DarkRed, text=white]{S}\hspace{4pt}\encircle[fill=DarkRed, text=white]{P}  & \encircle[fill=myBlue, text=white]{A}& & \ding{52}&\\
 BEATS \citep{sun2024beats}& arXiv'24& GSM8K/MATH/SVAMP/SimulEq/NumGLUE& \encircle[fill=DarkRed, text=white]{S}\hspace{4pt}\encircle[fill=DarkRed, text=white]{P}   & \encircle[fill=myBlue, text=white]{A}& & \ding{52}&\\
 MindStar \citep{kang2024mindstar}& arXiv'24& GSM8K/MATH& \encircle[fill=DarkRed, text=white]{S}\hspace{4pt}\encircle[fill=DarkRed, text=white]{P}   & \encircle[fill=myBlue, text=white]{A}& & \ding{52}&\\
 UMM \citep{zhang2024unconstrained}& arXiv'24& MMLU/GSM8K-COT/GSM8K-Coding/MATH-COT/MATH-Coding/HumanEval/InfiBench & \encircle[fill=DarkRed, text=white]{S}\hspace{4pt}\encircle[fill=DarkRed, text=white]{P}   & \encircle[fill=myBlue, text=white]{A}& & \ding{52}&\\
 STIC \citep{deng2024enhancing} \FiveStarConvex & arXiv'24& ScienceQA/TextVQA/ChartQA/LLaVA-Bench/MMBench/MM-Vet/MathVista& \encircle[fill=DarkRed, text=white]{S}  & \encircle[fill=myOrangev2, text=white]{M}& & \ding{52}&\\
 SPMWPs \citep{zhang2023interpretable}& ACL'23& GSM8K& \encircle[fill=DarkRed, text=white]{S}  & \encircle[fill=myBlue, text=white]{A}& & \ding{52}&\\
 CoRe \citep{zhu2022solving}& ACL'23& GSM8K/ASDiv-A/SingleOp/SinlgeEq/MultiArith& \encircle[fill=DarkRed, text=white]{S}  & \encircle[fill=myBlue, text=white]{A}& & \ding{52}&\\
 TabMWP \citep{lu2022dynamic}& ICLR'23& TabMWP& \encircle[fill=DarkRed, text=white]{S}& \encircle[fill=myBlue, text=white]{A}\hspace{4pt}\encircle[fill=myYellowv2, text=white]{D}& & \ding{52}&\\
 Chameleon \citep{lu2024chameleon} \FiveStarConvex & NeurIPS'23& ScienceQA/TabMWP& \encircle[fill=DarkRed, text=white]{S}& \encircle[fill=myBlue, text=white]{A}\hspace{4pt}\encircle[fill=myYellowv2, text=white]{D}& & &\ding{52}\\
 ATHENA \citep{kim2023athena}& EMNLP'23& MAWPS/ASDivA/Math23k/SVAMP/Un-biasedMWP& \encircle[fill=DarkRed, text=white]{S}\hspace{4pt}\encircle[fill=DarkRed, text=white]{P}& \encircle[fill=myBlue, text=white]{A}& & \ding{52}&\\
 UniMath \citep{liang2023unimath} \FiveStarConvex & EMNLP'23& SVAMP/GeoQA/TabMWP/MathQA/UniGeo-Proving& \encircle[fill=DarkRed, text=white]{S}\hspace{4pt}\encircle[fill=DarkRed, text=white]{P}& \encircle[fill=myBlue, text=white]{A}\hspace{4pt}\encircle[fill=myYellowv2, text=white]{D}& & \ding{52}&\\
 Jiuzhang 2.0 \citep{zhao2023jiuzhang}& KDD'23& MCQ/BFQ/CAG/BAG/KPC/QRC/JCAG/JBAG& \encircle[fill=DarkRed, text=white]{S}& \encircle[fill=myBlue, text=white]{A}& & \ding{52}&\\
 TCDP \citep{qin2023template}& TNNLS'23& Math23k/CM17K& \encircle[fill=DarkRed, text=white]{S}& \encircle[fill=myBlue, text=white]{A}& & \ding{52}&\\
 UniGeo \citep{chen2022unigeo} \FiveStarConvex & EMNLP'22& GeoQA/UniGeo & \encircle[fill=DarkRed, text=white]{S}\hspace{4pt}\encircle[fill=DarkRed, text=white]{P}& \encircle[fill=MyGreenv2, text=white]{G}& & \ding{52}&\\
 LogicSolver \citep{yang2022logicsolver}& EMNLP Findings'22& InterMWP/Math23K& \encircle[fill=DarkRed, text=white]{S}& \encircle[fill=myBlue, text=white]{A}& \ding{52}& \ding{52}&\\
 Jiuzhang \citep{zhao2022jiuzhang}& KDD'22& KPC/QRC/QAM/SQR/QAR/MCQ/BFQ/CAG/BAG& \encircle[fill=DarkRed, text=white]{S}& \encircle[fill=myBlue, text=white]{A}& & \ding{52}&\\
 MWP-BERT \citep{liang2021mwp}& NAACL'22& Math23k/MathQA/Ape-210k& \encircle[fill=DarkRed, text=white]{S}& \encircle[fill=myBlue, text=white]{A}& & \ding{52}&\\
 Inter-GPS \citep{lu2021inter} \FiveStarConvex & ACL'21& Geometry3K/GEOS& \encircle[fill=DarkRed, text=white]{S}& \encircle[fill=MyGreenv2, text=white]{G}& & \ding{52}&\\
 \bottomrule
    \end{tabular}
}

\caption{\textbf{Overview of LLM-based methods for mathematical reasoning}.  \FiveStarConvex refers to those specifically designed to tackle the multimodal mathematical setting. Different colors indicate different types for the following columns:
\\ \textbf{Task}: \encircle[fill=DarkRed, text=white]{S} = Problem-\underline{S}olving, \encircle[fill=DarkRed, text=white]{D} = Error \underline{D}etection, \encircle[fill=DarkRed, text=white]{P} = \underline{P}roving, \encircle[fill=DarkRed, text=white]{O} = \underline{O}thers
\\ \textbf{Scope}: \encircle[fill=MyGreenv2, text=white]{G} = \underline{G}eometry, \encircle[fill=myBlue, text=white]{A} = \underline{A}lgebra, \encircle[fill=myYellowv2, text=white]{D} = \underline{D}iagram, \encircle[fill=myOrangev2, text=white]{M} = General \underline{M}ath
}

\label{tab:method}
\end{table*}

\section{More Details of Challenges}
\label{app:challenge_details}

\subsection{Discussion of Data Bottlenecks}
\label{app:data_bottleneck}
We dive into the three bottlenecks of multimodal mathematical datasets as follows.

\ding{182} \textbf{Bottleneck in Data Quality:}

\begin{enumerate}
    \item \textbf{Labeling Noise and Modality Alignment}: Multimodal math problems often involve complex associations between text, formulas, and charts. Mismatches between text descriptions and images/formulas (\textit{e.g., }incorrect axis labels, contradictions between geometry figures and problem statements) can severely impair the model's ability to understand cross-modal relationships.
    \item \textbf{Lack of Deep Annotation for Problem Solving Process}: Most datasets only provide final answers, lacking intermediate steps such as algebraic transformations or construction of geometric auxiliary lines, making it difficult for models to learn the mathematical thinking chain (Chain-of-Thought).
\end{enumerate}

\ding{183} \textbf{Bottleneck in Data Diversity:}

\begin{enumerate}
    \item Limited Coverage of Problem Types and Scenarios: Existing datasets are mostly focused on basic math areas (\textit{e.g., }algebraic equations, simple geometry) and insufficiently cover higher-level math (\textit{e.g., }topology, discrete mathematics) or real-world scenarios (\textit{e.g.,} physics modeling, financial calculations).
    \item Monotony in Multimodal Combination Patterns: Modal interactions are often simple concatenations (\textit{e.g.,} text + static charts) without dynamic interactions (\textit{e.g.,} scalable geometric figures), or multi-step cross-modal reasoning (\textit{e.g.,} generating charts from text descriptions and then solving problems).
\end{enumerate}

\ding{184} \textbf{Bottleneck in Data Scale:}

\begin{enumerate}
    \item \textbf{High Cost of High-Quality Data Acquisition}: Mathematical problems need to be designed by experts and ensure multimodal consistency, which leads to long production cycles and high costs. Additionally, there is data scarcity for long-tail problems (\textit{e.g.,} niche branches of mathematics), which cannot be supplemented by scraping existing resources (\textit{e.g.,} textbooks, online question banks).
    \item \textbf{Imbalance in Modal Data Volumes}: Text data volumes far exceed those of image/symbol modalities, leading to models' insufficient feature extraction capability for non-text modalities.
\end{enumerate}

\ding{185} Based on recent trends in the latest works, we further propose the following \textbf{actionable suggestions to address these dataset bottlenecks}:

\begin{enumerate}
    \item \textbf{Innovation in Data Generation Techniques}: Combine formal mathematical engines (\textit{e.g.,} Lean, Coq) to generate verifiable reasoning steps, use programmatic rendering tools (\textit{e.g.,} TikZ, GeoGebra) to automatically generate precise charts, and design semi-automated annotation pipelines that reduce manual labor through large models generating drafts and experts refining them.
    \item \textbf{Diversity Enhancement Strategies}: Construct interdisciplinary, cross-cultural benchmark datasets (\textit{e.g.,} math-physics cross-domain problems), utilize crowdsourcing platforms to collect real-world scenario problems, and explore controllable data augmentation techniques, such as rule-based problem deformation (\textit{e.g.,} modifying parameters or replacing chart elements).
    \item \textbf{Scaling and Resource Integration}: Encourage collaborative dataset creation within the academic community (similar to ProofWiki), integrate existing educational resources (\textit{e.g.,} Khan Academy video-text analysis), and use synthetic data to fill long-tail gaps while improving model robustness to synthetic noise through adversarial training.
\end{enumerate}

\subsection{Limited Domain Generalization in Multimodal Contexts}
\label{app:limited_domain_generalization}
We further discuss the challenge of \textit{limited domain generalization} in multimodal contexts through the perspective of the methodology paradigm.
\begin{enumerate}
    \item \textbf{LLM as Enhancer}: Generate mixed-domain problems (\textit{e.g.,} combining algebraic equations with geometric figures) to force the model to learn cross-domain associations. Explicitly add domain labels (\textit{e.g.,} "spatial reasoning" label for geometry problems) to guide the model in distinguishing domain-specific features. The limitation of this paradigm is that enhanced data may lack the real-world complexity of domain intersections.
    \item \textbf{LLM as Reasoner}: Fine-tune the model separately for different mathematical domains (\textit{e.g., }algebra, geometry) to learn domain-specific visual patterns (\textit{e.g.,} encoding geometric properties in figures). Use domain-specific few-shot examples (\textit{e.g.,} providing figure-text associations in geometry) to guide the model in switching reasoning modes. The limitation is that the model’s capacity may be limited, making it difficult to master multiple significantly different domains simultaneously (\textit{e.g.,} switching from algebraic symbol manipulation to geometric spatial reasoning).
    \item \textbf{LLM as Planner}: Based on the problem domain (\textit{e.g.,} detecting the "triangle" keyword), call specialized tools (\textit{e.g.,} geometric theorem prover). For composite problems (\textit{e.g.,} algebraic-geometry equations), coordinate symbolic computation tools (\textit{e.g.,} Mathematica) and graphical reasoning tools (\textit{e.g.,} GeoGebra). The limitation is that domain boundary issues (\textit{e.g.,} math word problems requiring commonsense reasoning) may fail to route to the appropriate tools.
\end{enumerate}

\subsection{Error Feedback Limitations in Multimodal Contexts}
\label{app:error_feedback_limitation}
We further discuss the challenge of \textit{error feedback limitations} in multimodal contexts through the perspective of the methodology paradigm.
\begin{enumerate}
    \item \textbf{LLM as Enhancer}: Inject cross-modal errors (e.g., plot errors in function curves while the text description is correct) to train the model to detect contradictions. The limitation is that labeling error types is costly and it’s difficult to cover all long-tail errors.
    \item \textbf{LLM as Reasoner}: Decompose reasoning into "computation-logic-conclusion" stages and cross-check text derivations with graphical information (\textit{e.g.,} verify function extrema calculations using coordinates in the image). The limitation is that self-doubt relies on the model’s prior knowledge of error types, potentially missing rare error patterns in the training data.
    \item \textbf{LLM as Planner}: Use OCR tools to extract symbols from figures and compare them with the text description to detect misunderstandings. The limitation is that tool invocation delays affect real-time performance, and some errors require manually defined detection rules.
\end{enumerate}

\subsection{How Test-Time Scaling Techniques Handle Other Challenges}
\label{app:test_time_scaling}
We believe that test-time scaling techniques \cite{xu2025towards,li2025search,besta2025reasoning,muennighoff2025s1,chen2024expanding,chen2025rethinking,hochlehnert2025sober,li2025system} can also help handle other challenges discussed in Section \ref{sec:challenge}, especially the following three challenges.

\ding{182} \textbf{Insufficient Visual Reasoning:}

\begin{enumerate}
    \item \textbf{Enhancement of Multimodal Reasoning Chains}: During reasoning, generate multi-step visual-symbol joint inference paths. For example, using CoT prompts to guide the model in decomposing geometric shapes into angle, side length, and other symbolic constraints, and then calling a geometry solver to validate spatial relationships \cite{luo2025ursa,deng2024r,wang2025multimodal}.
    \item \textbf{Visual-Symbol Alignment Verification}: Use Best-of-N sampling to generate multiple candidate diagram parsing results and call external OCR tools or geometry validators (\textit{e.g.,} GeoGebra) to detect visual-text consistency and filter out erroneous explanations \cite{wu2025advanced,qi2025large,liu2025enhancing,ranaldi2025improving}.
    \item  \textbf{Limitations}: Parsing complex visual details (\textit{e.g.,} topological structures) depends on the pretrained visual encoder's capabilities. If the training data coverage is insufficient, test-time strategies may not be able to compensate \cite{ke2025survey,chen2025towards}.
\end{enumerate}

\ding{183} \textbf{Limited Domain Generalization:}

\begin{enumerate}
    \item Dynamic Domain Routing: Use Beam Search Process Reward Model (PRM) to select domain-specific inference paths based on problem types (\textit{e.g.,} detecting the “triangle” keyword and choosing between algebra solvers or geometry theorem provers) \cite{zhao2025genprm,zeng2025versaprm}.
    \item Meta-learning Optimization: Fine-tune the model on a small number of domain-specific samples via Test-Time Training (TTT) to quickly adapt to new domains (\textit{e.g.,} probability and statistics problems)
    \item Limitations: Problems with blurred domain boundaries (\textit{e.g., }math application problems involving common sense reasoning) may fail due to routing errors.
\end{enumerate}

\ding{184} \textbf{Error Feedback Limitations:}

\begin{enumerate}
    \item Process Supervision Reinforcement: Use PRM to validate each step of reasoning in real-time. If an error is detected (\textit{e.g.,} misuse of integration symbols), backtrack and correct the path; combine Self-Consistency by generating multiple inference paths and selecting the one with no contradictions via majority voting \cite{wang2025visualprm,zhang2025lessons,song2025prmbench,cui2025process,hu2025coarse,ji2025survey,zhang2025r1}.
    \item Limitations: The reliability of PRM depends on the coverage of error types in the training data. Long-tail errors such as rare symbol confusions may be overlooked.
\end{enumerate}

In summary, combining the flexibility of test-time scaling with the specialization of multimodal tools can help mitigate the core challenges in multimodal mathematical reasoning. However, \textbf{it is crucial to balance computational efficiency and accuracy}.

\end{document}